\documentclass{article} 
\usepackage{iclr2026_conference,times}

\usepackage{amsmath,amsfonts,bm}

\def\eqref#1{equation~\ref{#1}}

\def\1{\bm{1}}

\DeclareMathAlphabet{\mathsfit}{\encodingdefault}{\sfdefault}{m}{sl}
\SetMathAlphabet{\mathsfit}{bold}{\encodingdefault}{\sfdefault}{bx}{n}

\usepackage{hyperref}
\usepackage{url}
\usepackage[pdftex]{graphicx} 
\usepackage{makecell} 
\usepackage{subcaption} 
\usepackage{float} 
\usepackage{booktabs}  

\title{Healthy LLMs? Benchmarking LLM Knowledge of UK Government Public Health Information}

\author{
\textbf{Joshua Harris$^\dagger$, Fan Grayson, Felix Feldman, Timothy Laurence,}\\
\textbf{Toby Nonnenmacher, Oliver Higgins, Leo Loman, Selina Patel, Thomas Finnie,}\\
\textbf{Samuel Collins, Michael Borowitz}\\
UK Health Security Agency (UKHSA)\\
}

\iclrfinalcopy 
\begin{document}

\maketitle
\lhead{Preprint.} 

\begin{abstract}
As Large Language Models (LLMs) become widely accessible, a detailed understanding of their knowledge within specific domains becomes necessary for successful real world use. This is particularly critical in the domains of medicine and public health, where failure to retrieve relevant, accurate, and current information could significantly impact UK residents. However, while there are a number of LLM benchmarks in the medical domain, currently little is known about LLM knowledge within the field of public health. To address this issue, this paper introduces a new benchmark, PubHealthBench, with over 8000 questions for evaluating LLMs' Multiple Choice Question Answering (MCQA) and free form responses to public health queries. To create PubHealthBench we extract free text from 687 current UK government guidance documents and implement an automated pipeline for generating MCQA samples. Assessing 24 LLMs on PubHealthBench we find the latest proprietary LLMs (GPT-4.5, GPT-4.1 and o1) have a high degree of knowledge, achieving over 90\% accuracy in the MCQA setup, and outperform humans with cursory search engine use. However, in the free form setup we see lower performance with no model scoring over 75\%. Therefore, while there are promising signs that state of the art (SOTA) LLMs are an increasingly accurate source of public health information, additional safeguards or tools may still be needed when providing free form responses.
\renewcommand*{\thefootnote}{\fnsymbol{footnote}}
\footnotetext[2]{joshua.harris@ukhsa.gov.uk}
\renewcommand*{\thefootnote}{\arabic{footnote}}
\end{abstract}

\begin{figure}[ht]
  \centering
  \begin{minipage}[t]{0.45\linewidth}
    \centering
    \includegraphics[width=\linewidth]{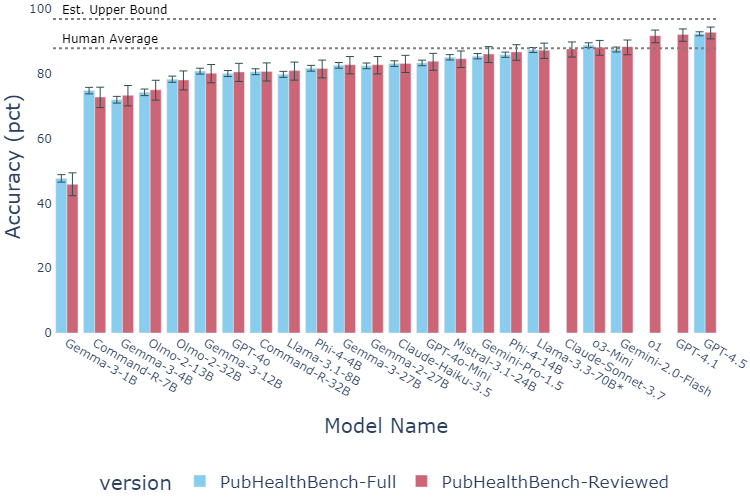}
  \end{minipage}
    \begin{minipage}[t]{0.45\linewidth}
    \centering
    \includegraphics[width=\linewidth]{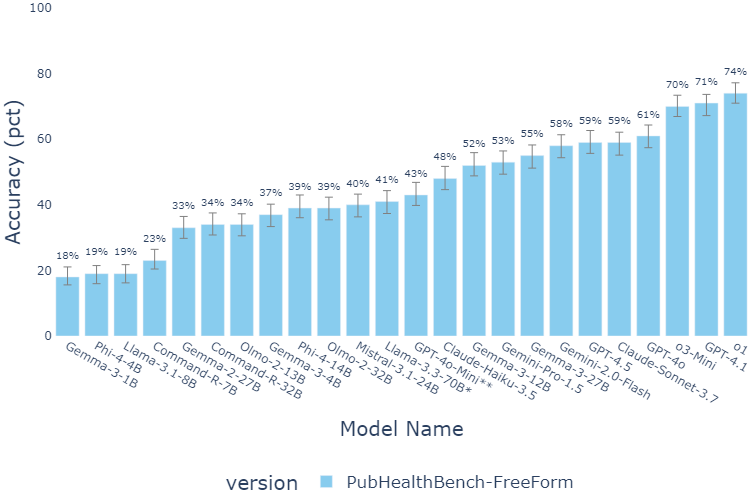}
  \end{minipage}
  \caption{(left) PubHealthBench Full and Reviewed model accuracy, (right) PubHealthBench-FreeForm model accuracy. 95\% Wilson CI. *LLM used to generate benchmark, **Judge LLM. }
  \label{fig:results_bar}
\end{figure}

\section{Introduction}

Public health guidance represents an important source of information for UK residents and experts to inform personal, professional, and clinical decision making. The release of highly capable Large Language Models (LLMs)~\citep{minaee2025largelanguagemodelssurvey}, and particularly chatbots~\citep{chatgpt_chatbot}, could represent a significant shift in how public health guidance is retrieved, analysed, and disseminated. This in turn raises significant opportunities and risks for public health institutions both internally and when engaging the public.

Whilst LLMs often undergo a broad range of evaluations during development~\citep{grattafiori2024llama3herdmodels, gemini_2_5, o3_mini, claude_3_7}, and there are a number of existing benchmarks in the medical domain, there is currently an important gap in  field of public health, with no comprehensive LLM benchmarks covering this domain, including for existing UK Government guidance.  Furthermore, due to guidance undergoing regular revisions, and differing guidance being issued across institutions and geographies, accurate up to date knowledge of UK public health guidance may be particularly challenging for LLM systems. Therefore, as recently observed for BBC news stories~\citep{bbcaiassitantsarticle}, there is a risk that LLM based applications and chatbots generate hallucinations~\citep{Huang_2025} or incomplete information regarding UK public health advice. This in turn could have a significant impact on the public. These risks, combined with the increasing desire within the UK Government to incorporate LLMs into existing real world processes~\citep{gov_ai_op, ukhsa_ai_board}, means comprehensive evaluations of LLMs' understanding of UK public health guidance are needed. 

In this paper we introduce a new dataset, Multiple-Choice Question Answering (MCQA) benchmark, and free form response benchmark for assessing LLMs' knowledge across a broad range of UK public health guidance.\footnote{The guidance included can cover the entire UK or individual constituent countries, most documents primarily relate to English public health guidance. We also only included English language documents.} Specifically our contributions include:

    \textbf{PubHealthBench a fully grounded MCQA benchmark} - We collect, extract, markdown format, and chunk information from over 500 publicly available UK Health Security Agency (UKHSA) PDF and HTML documents from the UK Government website (gov.uk).\footnote{GOV.UK reuse policy: \url{https://www.gov.uk/help/reuse-govuk-content}} We implement an automated MCQA generation and validation pipeline grounded in the extracted guidance source text. This enables us to generate a new benchmark with over 8000 MCQA questions to test LLM knowledge across a broad range of current guidance. We provide results for both the full benchmark (PubHealthBench-Full) and a manually reviewed subset (PubHealthBench-Reviewed).
    
    \textbf{PubHealthBench-FreeForm} - To assess LLMs in a more realistic real-world setting we also implement a free form response benchmark using the questions from the manually reviewed subset. By utilising the fact that every question can be linked back to the original source chunk and document, we implement a grounded LLM judge to assess responses.  
    
    \textbf{Initial LLM evaluations} - We evaluate 24 private and open-weight LLMs on this new public health benchmark. Given our focus on assessing knowledge, we primarily focus on SOTA non-reasoning models, but also include some leading reasoning models for comparison.
    
    \textbf{Manual human expert review and human baseline} - To quality assure the benchmark and establish an upper bound for performance, human experts manually reviewed a random sample of 800 MCQA examples (approx. 10\% of the benchmark). Furthermore, in order to compare to human performance, 5 humans also took sample tests (total 600 MCQA examples) to establish a human baseline with the use of search engines. This provides an initial indication of the boundary at which LLMs become similarly accurate to a cursory search by non-expert humans.

\begin{figure}[ht]
    \centering
    \includegraphics[width=\linewidth]{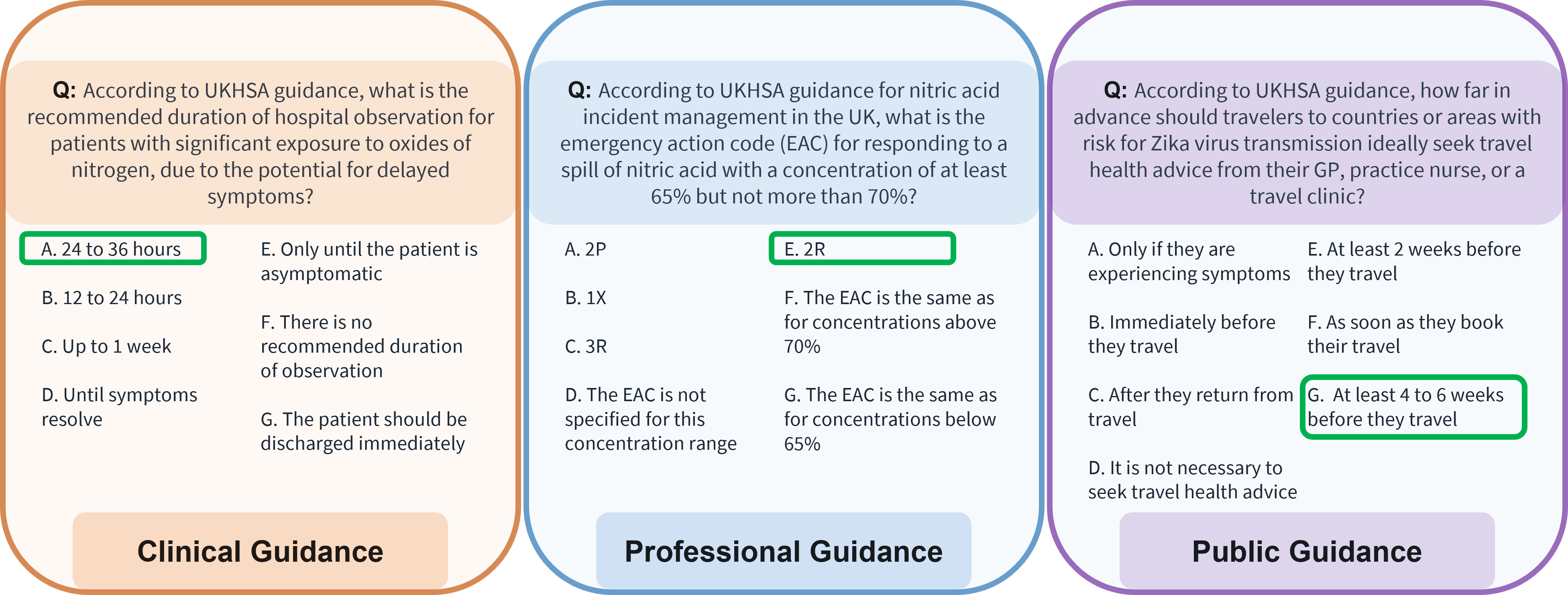} 
    \caption{Example PubHealthBench MCQA benchmark questions.} \label{fig:example_questions}
\end{figure}

\section{Related work}

\subsection{MCQA LLM evaluations}

Using Multiple-Choice Question Answering (MCQA) to assess the knowledge of LLMs is well established in the literature. Earlier work evaluating LLM knowledge in specific domains often used existing MCQA human assessments and exams~\citep{openai2023gpt4, bommarito2023gptknowledgeworkerzeroshot, jin2020diseasedoespatienthave, pal2022medmcqalargescalemultisubject}, for example in the medical domain using evaluations from the US Medical Licensing
Examination (USMLE)~\citep{chatgptUSMLE, singhal2023large_usmle}.

Broader evaluations of LLM knowledge and capabilities have also been introduced.  One of the most widely adopted being the Massive Multitask Language Understanding (MMLU) benchmark~\citep{hendrycks2021measuringmassivemultitasklanguage}. More recently the MMLU-Pro benchmark by~\citet{wang2024mmluprorobustchallengingmultitask}, updated the original MMLU evaluations for errors~\citep{gema2024mmlu} and increased the question difficulty. New domain specific MCQA LLM evaluations have also been created, such as the GPQA~\citep{rein2023gpqagraduatelevelgoogleproofqa} and ARC~\citep{clark2018thinksolvedquestionanswering} benchmarks within science.

\subsection{Synthetic MCQA datasets}

There is increasing research investigating automated approaches to MCQA generation grounded in existing corpora~\citep{shashidhar2025yourbencheasycustomevaluation, guinet2024automatedevaluationretrievalaugmentedlanguage, ghazaryan2024syndarinsynthesisingdatasetsautomated}. To develop MMLU-Pro, ~\citet{wang2024mmluprorobustchallengingmultitask} use GPT4-Turbo to generate answer options, as well as to augment the option lists found in other LLM benchmarks (primarily MMLU) with additional distractors. Similarly a combined human-LLM pipeline is used to construct the recent SuperGPQA benchmark~\citep{pteam2025supergpqascalingllmevaluation} and in \citet{asiedu2025contextualevaluationlargelanguage}'s work evaluating LLMs for tropical and infectious diseases. Concurrent work by~\citet{shashidhar2025yourbencheasycustomevaluation} on the YourBench framework goes further and provides a fully automated approach to generating MMLU style evaluations from new document corpora.

\subsection{Public health evaluations} \label{sec:pipeline}

While there exists a range of benchmarks within clinical medicine ~\citep{jin2020diseasedoespatienthave, pal2022medmcqalargescalemultisubject, arora2025healthbenchevaluatinglargelanguage}, the public health domain, which focuses on areas such as prevention, environmental hazards, community interventions, and managing infectious disease outbreaks, does not have an existing LLM benchmark. In public health, the closest evaluations to those in this paper were performed by~\citet{davies2024chatgpt}, which assessed ChatGPT 3.5's open-ended responses to the UK Faculty of Public Health’s Diplomate exam (DFPH) Paper 1 questions. ChatGPT was evaluated on 119 questions double marked by DFPH examiners.~\citet{ayers2023evaluating} conduct similar evaluations using 23 questions across 4 public health domains. Finally, work by~\citet{harris2024evaluatinglargelanguagemodels} evaluated LLMs on classification and extraction tasks using public health guidance free text.

\begin{figure}[ht]
    \centering
    \includegraphics[width=\linewidth]{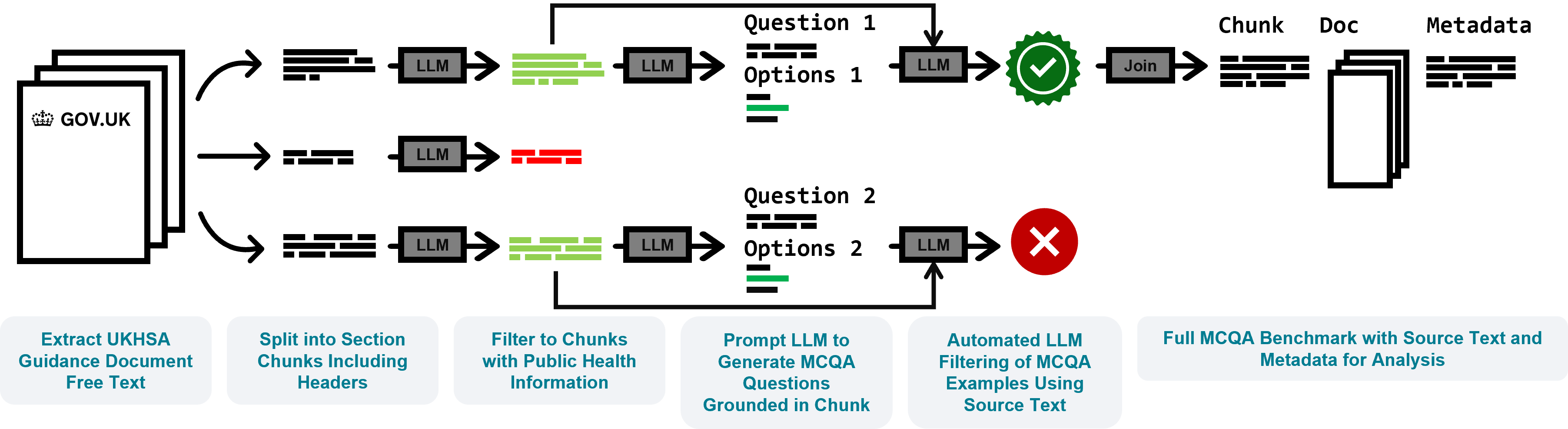} 
    \caption{Overview of MCQA generation pipeline.} \label{fig:pipeline_fig}
\end{figure}
\section{Methods}

To generate our MCQA benchmark we develop an automated pipeline (Figure~\ref{fig:pipeline_fig}) to extract free text from documents, chunk it into sections, generate MCQA samples, and filter to a high quality subset. We focus on an automated approach for three reasons: (1) generating thousands of MCQA samples manually is highly time consuming, (2) public health guidance is frequently revised and so any approach needs to be amenable to regular updates, and (3) it allows us to easily extend our benchmarking to additional knowledge bases in the future.

\subsection{UK public health information}

Public health covers a very broad range of topics from biosecurity to tackling health inequalities. To assess LLMs' knowledge of UK guidance across these areas we collect a large corpus of 1,150 current UK Government guidance documents from the UK Government website (gov.uk) in HTML and PDF formats. Using publicly available source documents is necessary for making the benchmark applicable to the real-world and so we assume parts of the document corpus are likely to be included in LLM pre-training datasets. However, by synthetically generating all questions and answer options we can be confident the benchmark was not in model training datasets.

\subsection{Pre-processing and chunking}

HTML documents are pre-processed and converted into markdown format. PDF document extraction is more challenging. Therefore, we use a two stage pipeline to achieve the requisite performance on PDF documents. We first extract the raw text from the PDFs using existing tools. We then use OpenAI's GPT-4o-mini vision LLM via the API, prompting the model to extract the text from the image (including markdown headers). For each page individually we pass: an image of the PDF page, the raw markdown text extracted using existing tools for that page at the first step, and the header hierarchy. We then split the documents into 20,488 smaller section chunks based on the markdown headers of each document, and include the hierarchy of higher level headers into every chunk to ensure relevant wider context and document structure is available.

\subsection{Question generation}

Guidance documents often contain significant background information and operational details that do not directly relate to UK public health recommendations. Therefore, in order to generate relevant questions we first use an LLM to classify each chunk into whether it contains a public health recommendation and filter out any chunks that do not. We also filter out chunks exceeding \textasciitilde{2000} words. This reduces the total number of chunks to 7,946, which form the source material for our MCQA generation.

We then use an LLM (Llama-3.3-70bn-Instruct) to generate two multiple choice questions per chunk in the standard MCQA format (Figure~\ref{fig:example_questions}), with: a single question, single correct answer option, and six incorrect distractor options per question. We use a one-shot with Chain of Thought (CoT) prompt instructing the LLM to output its final answer as a JSON (see Appendix~\ref{supp:generation_prompts}). To ensure the LLM has the required context it is also provided with the text chunks that appear either side of the target chunk in the document (whether or not these contain recommendations). We run this pipeline across all of the filtered text chunks generating 15,666 correctly formatted candidate MCQA samples.

\subsection{Automated MCQA error detection and sampling}

To check consistency with the source text and improve the quality of our question set, we use LLMs to filter potentially invalid questions. For full details of the approach and evaluation see Appendix~\ref{supp:auto_error}. We select the Llama-3-70bn-Instruct model for this error detection step and filter the 15,666 candidate MCQA questions down to 14,440 that were not flagged as containing potential errors. Finally, we remove questions relating to documents that contain guidance that has been withdrawn, and balance our dataset between HTML and PDF documents to ensure a more even representation of topics (as PDF documents were often longer). We retain the remaining approximately 4,000 unused questions from PDF source documents as a potential internal hold-out set. 

\subsection{Final MCQA benchmark dataset}

\begin{figure}[ht]
  \centering
  \begin{minipage}[t]{0.49\linewidth}
    \centering
    \includegraphics[width=\linewidth]{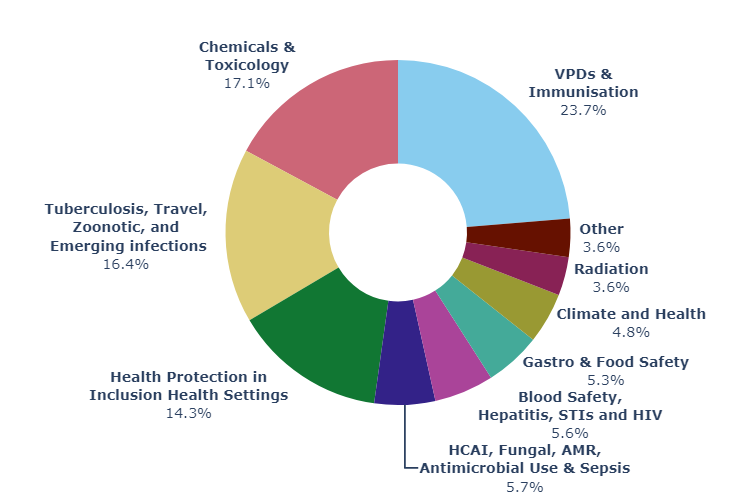}
  \end{minipage}\hfill
  \begin{minipage}[t]{0.49\linewidth}
    \centering
    \includegraphics[width=\linewidth]{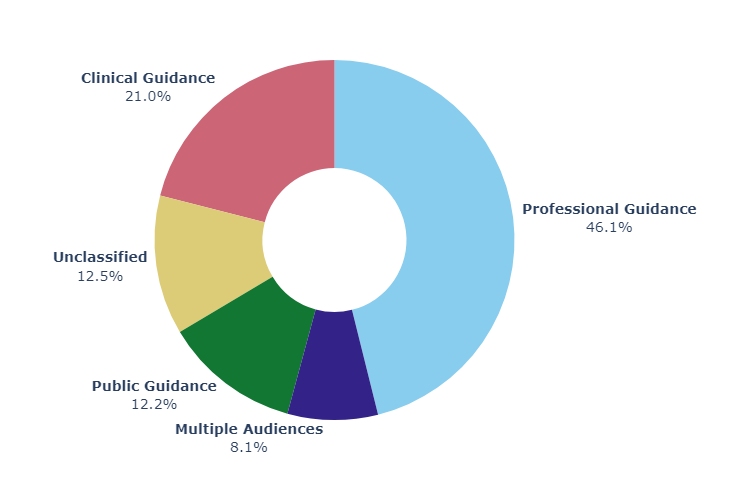}
  \end{minipage}
  \caption{PubHealthBench questions by guidance topic area and guidance audience.}
  \label{fig:combined}
\end{figure}

 Overall, the final public benchmark (PubHealthBench) consists of 8,090 MCQA questions covering public, clinical, and professional guidance across 10 public health topic areas, and 352 guidance areas, sourced from 687 documents containing UK Government public health information. Figure~\ref{fig:combined} provides a breakdown of the benchmark by topic area and the intended audience for the guidance.

\begin{figure}[ht]
    \centering
    \includegraphics[width=\linewidth]{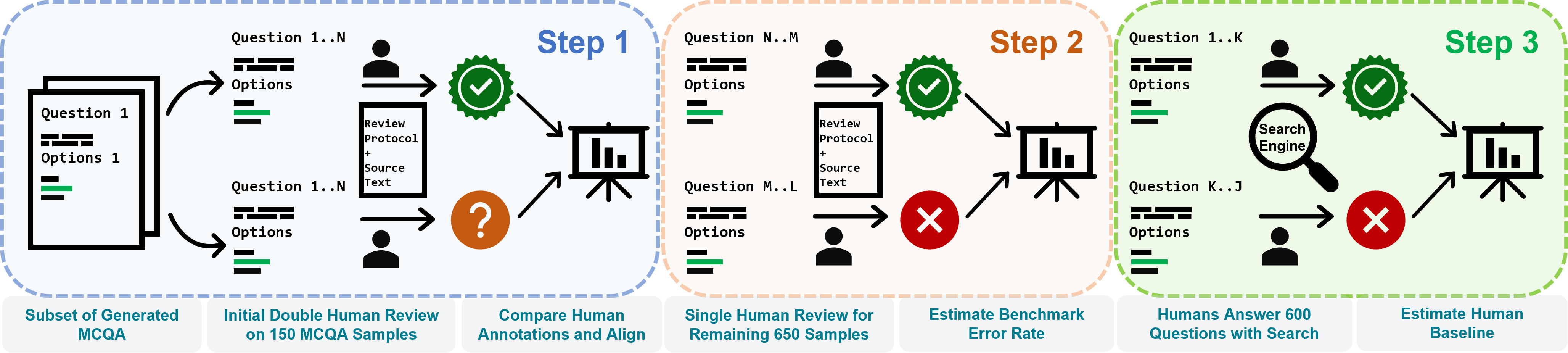} 
    \caption{Overview of benchmark review steps.} \label{fig:bench_validation}
\end{figure}

\subsection{Human expert quality assurance}\label{sec:human_manual_review}

To quality assure the benchmark and estimate the underlying rate of invalid questions, we manually review a random sample of 800 questions (c.10\% of the benchmark). Based on this manual review we estimate the rate of ambiguous or invalid questions in the full benchmark to be approximately 5.5\% (4.1\%-7.3\%, 95\% Wilson score CI) in the final dataset. See Appendix~\ref{sec:supp_human_annotation} for annotation details.

However, the majority of questions identified as invalid related to situations where one of the distractor options could be considered an equally good answer to the specified correct answer. Therefore, whilst these questions were deemed ambiguous, the correct answer should remain one of the most likely guesses for LLMs. Accounting for this random guessing over usually two correct answers, we expect the upper bound score on this benchmark to be approximately 97\%. This is supported by the observed model performance on questions classified as invalid, as discussed in Section~\ref{sec:results}.

\section{Model evaluation approach}

To understand the level of knowledge current open-weight and proprietary LLMs have about UK Government public health guidance, we assess 24 models on PubHealthBench, including: GPT-4.5~\citep{gpt_4_5}, Claude-Sonnet-3.7~\citep{claude_3_7}, Gemma-3~\citep{gemma_2025}, o1~\citep{o1_sys_card}, Phi-4~\citep{abdin2024phi4technicalreport}, OLMo-2~\citep{olmo20252olmo2furious}, and Llama-3.3~\citep{grattafiori2024llama3herdmodels}. 

\subsection{Benchmark subsets}

We report overall results for three subsets of PubHealthBench:

\textbf{PubHealthBench-Full} - the full MCQA benchmark, providing the broadest assessment of LLM capabilities. This enables us to provide results across granular topic areas within public health.

\textbf{PubHealthBench-Reviewed} - the random test subset of 760 MCQA questions that have been manually reviewed by human experts, we report results both including and excluding questions classified as ambiguous or invalid so as to make the results comparable to the full benchmark. For the most expensive models we run this subset instead of the full benchmark for cost reasons.

\textbf{PubHealthBench-FreeForm} - the same manually reviewed subset as in \textit{PubHealthBench-Reviewed} but only asking the question (without multiple choice options) and allowing for open-ended free form responses, similar to how a chatbot may respond in real world uses cases. We use a grounded LLM judge to assess whether the free text answer is consistent with the source material.

\subsection{Human baseline}

In addition to estimating the theoretical upper bound (97\%), we also provide baseline human performance for comparability (Figure~\ref{fig:bench_validation}). Human test takers answered the MCQA samples \textit{with} access to search engines but \textit{without} any LLM or AI enabled tools. Test takers were not trained public health specialists, and were encouraged to take no longer than 2 minutes per question. Under these conditions, humans scored 88\% on 600 questions, about 9 percentage points below the potential upper bound. We highlight that this result is meant to simulate a member of the public looking for relevant guidance relatively quickly. See Appendix~\ref{sec:supp_human_annotation} for further details on the human baseline setup.

\subsection{Experimental setup}

We closely follow the prompts and answer extraction used in the MMLU-Pro benchmark~\citep{wang2024mmluprorobustchallengingmultitask}. However, as with the human baseline, in our LLM evaluations we seek to replicate a similar query setup to that which might occur when interacting with a chatbot or within a simple LLM based application. Therefore, we focus on zero-shot prompting (see Appendix~\ref{supp:bench_prompts} for the prompt templates), and only use CoT when it is the default behavior, as for reasoning models. For comparability across models and to match the highest risk real world deployments, we do not allow any LLMs access to external tools (e.g search) or information repositories.

\subsubsection{Free form response evaluation}

Utilising the fact all questions are directly grounded in specific parts of the original source text we use an LLM as a Judge setup~\citep{zheng2023judgingllmasajudgemtbenchchatbot, gu2025surveyllmasajudge}  for the free form answer evaluation. We prompt a judge LLM (GPT-4o-Mini) with the question, ground truth answer, the LLM response, and six retrieved related chunks. The judge is asked to assess the response and provide a binary classification for whether it is consistent with the source text and ground truth MCQA answer. For full details see Appendix~\ref{supp:judge_setup}.

\section{Results}\label{sec:results}
\subsection{PubHealthBench - overall MCQA benchmark results}

\begin{table}
    \caption{PubHealthBench-Full - zero-shot accuracy for test set of 7929 questions, refusals included as incorrect responses, and bold indicates the highest score. *LLM used to generate benchmark.}
    \vspace{1em} 
    \label{tab:results_full}
    \centering
        \resizebox{\linewidth}{!}{%
    
\begin{tabular}{lccccccccccc}
\toprule
 & \makecell{Blood Safety,\\Hepatitis, STIs\\and HIV} & \makecell{Chemicals and\\Toxicology} & \makecell{Climate and\\Health} & \makecell{Gastro and\\Food Safety} & \makecell{HCAI, Fungal,\\ AMR,\\Antimicrobial\\Use and Sepsis} & \makecell{Health Protection\\in Inclusion\\Health Settings} & \makecell{Other} & \makecell{Radiation} & \makecell{Tuberculosis,\\Travel,\\Zoonotic, and\\ Emerging infections} & \makecell{VPDs and\\Immunisation} & \makecell{Overall} \\
Model Name &  &  &  &  &  &  &  &  &  &  &  \\
\midrule
GPT-4.5 & \textbf{90.9} & \textbf{91.1} & \textbf{97.4} & \textbf{90.3} & \textbf{94.6} & \textbf{94.0} & \textbf{91.3} & \textbf{89.9} & \textbf{91.8} & \textbf{93.0} & \textbf{92.5} \\
o3-Mini & 87.7 & 88.5 & 94.5 & 88.1 & 92.4 & 90.6 & 87.5 & 87.1 & 87.1 & 88.3 & 88.9 \\
Gemini-2.0-Flash & 84.7 & 86.1 & 95.3 & 86.0 & 88.2 & 90.6 & 87.1 & 88.5 & 86.3 & 87.4 & 87.7 \\
Llama-3.3-70B* & 86.5 & 86.6 & 92.4 & 85.5 & 87.9 & 90.9 & 86.1 & 87.5 & 85.3 & 87.2 & 87.4 \\
Phi-4-14B & 85.4 & 82.7 & 92.1 & 85.5 & 90.4 & 88.7 & 89.5 & 84.7 & 85.1 & 85.4 & 86.1 \\
Gemini-Pro-1.5 & 81.3 & 81.6 & 93.9 & 84.1 & 87.3 & 90.0 & 86.1 & 85.0 & 84.1 & 86.0 & 85.6 \\
Mistral-3.1-24B & 85.4 & 82.7 & 92.1 & 84.8 & 89.7 & 88.6 & 87.8 & 84.7 & 83.0 & 83.5 & 85.1 \\
GPT-4o-Mini & 83.1 & 80.1 & 91.6 & 81.9 & 88.6 & 88.1 & 86.4 & 82.9 & 79.9 & 82.8 & 83.5 \\
Claude-Haiku-3.5 & 82.4 & 80.5 & 92.4 & 80.3 & 86.2 & 87.6 & 86.4 & 86.1 & 81.1 & 81.5 & 83.2 \\
Gemma-3-27B & 84.5 & 79.7 & 91.6 & 80.3 & 83.9 & 87.4 & 84.3 & 80.5 & 80.1 & 82.0 & 82.7 \\
Gemma-2-27B & 84.2 & 79.3 & 91.3 & 82.9 & 83.9 & 86.9 & 84.3 & 80.5 & 80.4 & 81.3 & 82.6 \\
Phi-4-4B & 82.0 & 79.7 & 90.3 & 81.2 & 87.1 & 85.7 & 86.4 & 77.4 & 78.6 & 80.3 & 81.8 \\
Gemma-3-12B & 80.4 & 79.0 & 89.5 & 76.5 & 83.0 & 84.4 & 86.4 & 81.2 & 79.1 & 79.2 & 80.8 \\
Command-R-32B & 79.7 & 77.2 & 89.5 & 76.7 & 83.7 & 84.8 & 83.6 & 77.7 & 79.1 & 80.7 & 80.7 \\
GPT-4o & 79.5 & 77.4 & 89.2 & 79.1 & 80.8 & 85.9 & 87.5 & 79.8 & 79.6 & 76.7 & 80.2 \\
Llama-3.1-8B & 79.2 & 77.2 & 89.2 & 77.9 & 84.2 & 84.8 & 84.3 & 78.0 & 76.2 & 79.3 & 80.0 \\
Olmo-2-32B & 76.7 & 74.7 & 88.9 & 75.5 & 82.8 & 83.4 & 80.8 & 77.4 & 76.4 & 77.1 & 78.4 \\
Command-R-7B & 76.3 & 72.3 & 86.3 & 74.1 & 76.3 & 79.0 & 82.9 & 71.4 & 70.3 & 73.1 & 74.7 \\
Olmo-2-13B & 76.0 & 69.8 & 87.6 & 74.1 & 76.8 & 78.8 & 80.8 & 74.2 & 71.4 & 72.8 & 74.4 \\
Gemma-3-4B & 73.5 & 69.7 & 85.8 & 71.5 & 73.9 & 79.2 & 81.5 & 67.9 & 69.8 & 67.4 & 72.2 \\
Gemma-3-1B & 49.5 & 49.4 & 57.9 & 45.1 & 50.9 & 51.4 & 57.1 & 44.6 & 44.8 & 42.1 & 47.6 \\
\bottomrule
\end{tabular}

    }
\end{table}

For the 21 models run on the PubHealthBench-Full (Table~\ref{tab:results_full}) and the PubHealthBench-Reviewed subset (Table~\ref{tab:results_verified}), we find a very high correlation in overall accuracy between the two sets, with a correlation coefficient of over 0.99, a rank correlation of 0.98, and an average absolute score difference of under 1 percentage point. Therefore, we compare overall results directly for models only run on PubHealthBench-Reviewed for cost reasons. We provide tables with 95\% Wilson score confidence intervals in Appendix~\ref{supp:conf_int}. 

On MCQA format questions we find the latest proprietary LLMs perform very strongly, with the highest scoring models GPT-4.5, GPT-4.1\footnotemark[1], and o1\footnotemark[1], all achieving over 90\% accuracy, above the human baseline and nearing the benchmark's estimated upper bound.

\footnotetext[1]{Results on the PubHealthBench-Reviewed subset.}

Smaller open-weight models also show a reasonable degree of knowledge of public health guidance with most 5-15bn parameter models scoring above 75\%. However, recent "reasoning" models (o1 and o3-Mini) perform similarly to "non-reasoning" models with little additional benefit from the extra test time compute in the MCQA setting. As shown in Table~\ref{tab:results_verified}, due to the nature of the MCQA issues identified during manual review, we find models still achieve on average 60\% accuracy on MCQA questions labeled invalid. This adds additional support to our estimated upper bound for the benchmark of approximately 97\%.

\noindent
\begin{minipage}[t]{0.48\linewidth}
  \centering
  \captionof{table}{PubHealthBench-Full zero-shot accuracy by guidance type. *LLM used to generate benchmark.}
  \label{tab:results_full_type}
  \resizebox{\linewidth}{!}{
\begin{tabular}{lcccccc}
\toprule
 & \makecell{Clinical\\Guidance} & \makecell{Multiple\\Audiences} & \makecell{Professional\\Guidance} & \makecell{Public\\Guidance} & \makecell{Unclassified} & \makecell{Overall} \\
Model Name &  &  &  &  &  &  \\
\midrule
GPT-4.5 & \textbf{91.5} & \textbf{93.7} & \textbf{91.9} & \textbf{96.1} & \textbf{92.1} & \textbf{92.5} \\
o3-Mini & 85.9 & 91.7 & 88.7 & 92.8 & 88.9 & 88.9 \\
Gemini-2.0-Flash & 84.8 & 89.7 & 87.6 & 93.1 & 86.3 & 87.7 \\
Llama-3.3-70B* & 84.9 & 89.5 & 87.1 & 91.7 & 87.1 & 87.4 \\
Phi-4-14B & 84.5 & 88.1 & 85.5 & 90.5 & 85.0 & 86.1 \\
Gemini-Pro-1.5 & 82.5 & 90.0 & 84.8 & 90.6 & 85.9 & 85.6 \\
Mistral-3.1-24B & 81.4 & 87.2 & 84.9 & 90.1 & 86.0 & 85.1 \\
GPT-4o-Mini & 80.7 & 85.8 & 83.1 & 87.9 & 83.9 & 83.5 \\
Claude-Haiku-3.5 & 79.5 & 85.3 & 83.0 & 87.4 & 84.9 & 83.2 \\
Gemma-3-27B & 78.7 & 86.0 & 82.3 & 87.8 & 83.9 & 82.7 \\
Gemma-2-27B & 79.4 & 84.4 & 82.1 & 87.5 & 83.4 & 82.6 \\
Phi-4-4B & 78.3 & 84.4 & 81.8 & 85.4 & 82.5 & 81.8 \\
Gemma-3-12B & 76.9 & 83.2 & 80.7 & 87.1 & 80.2 & 80.8 \\
Command-R-32B & 77.7 & 83.2 & 79.5 & 86.9 & 82.6 & 80.7 \\
GPT-4o & 73.7 & 83.3 & 80.9 & 86.2 & 80.7 & 80.2 \\
Llama-3.1-8B & 76.8 & 82.3 & 80.1 & 82.6 & 81.1 & 80.0 \\
Olmo-2-32B & 74.6 & 81.8 & 78.1 & 84.9 & 77.4 & 78.4 \\
Command-R-7B & 70.1 & 73.6 & 75.6 & 79.3 & 75.4 & 74.7 \\
Olmo-2-13B & 69.6 & 77.2 & 74.2 & 81.2 & 74.9 & 74.4 \\
Gemma-3-4B & 64.7 & 74.2 & 73.1 & 77.7 & 74.7 & 72.2 \\
Gemma-3-1B & 40.1 & 45.2 & 50.3 & 48.7 & 50.5 & 47.6 \\
\bottomrule
\end{tabular}

}
  \vspace{1em}
\end{minipage}\hfill
\begin{minipage}[t]{0.48\linewidth}
  \centering
  \captionof{table}{PubHealthBench-Reviewed zero-shot accuracy by question and response type. *LLM used to generate benchmark, **Headline result.}
  \label{tab:results_verified}
  \resizebox{\linewidth}{!}{
\begin{tabular}{lcccc}
\toprule
 & \makecell{Exc. Refusals} & \makecell{Inc. Refusals**} & \makecell{Invalid MCQA} & \makecell{Valid MCQA} \\
Model Name &  &  &  &  \\
\midrule
GPT-4.5 & \textbf{92.9} & \textbf{92.9} & 71.4 & \textbf{94.2} \\
GPT-4.1 & 92.2 & 92.2 & \textbf{78.6} & 93.0 \\
o1 & 91.8 & 91.8 & 66.7 & 93.3 \\
Gemini-2.0-Flash & 88.5 & 88.4 & 61.9 & 90.0 \\
o3-Mini & 88.3 & 88.3 & 69.0 & 89.4 \\
Claude-Sonnet-3.7 & 92.4 & 87.8 & 59.5 & 89.4 \\
Llama-3.3-70B* & 87.4 & 87.4 & 61.9 & 88.9 \\
Phi-4-14B & 86.8 & 86.8 & 66.7 & 88.0 \\
Gemini-Pro-1.5 & 86.2 & 86.2 & 59.5 & 87.7 \\
Mistral-3.1-24B & 84.7 & 84.7 & 61.9 & 86.1 \\
GPT-4o-Mini & 83.9 & 83.9 & 52.4 & 85.8 \\
Claude-Haiku-3.5 & 83.3 & 83.3 & 57.1 & 84.8 \\
Gemma-3-27B & 82.9 & 82.9 & 59.5 & 84.3 \\
Gemma-2-27B & 82.9 & 82.9 & 54.8 & 84.5 \\
Phi-4-4B & 81.7 & 81.7 & 64.3 & 82.7 \\
Llama-3.1-8B & 81.1 & 81.1 & 57.1 & 82.5 \\
Command-R-32B & 80.8 & 80.8 & 59.5 & 82.0 \\
GPT-4o & 91.8 & 80.7 & 52.4 & 82.3 \\
Gemma-3-12B & 80.3 & 80.3 & 61.9 & 81.3 \\
Olmo-2-32B & 78.2 & 78.2 & 54.8 & 79.5 \\
Olmo-2-13B & 75.1 & 75.1 & 57.1 & 76.2 \\
Gemma-3-4B & 73.4 & 73.4 & 54.8 & 74.5 \\
Command-R-7B & 72.9 & 72.9 & 57.1 & 73.8 \\
Gemma-3-1B & 45.9 & 45.9 & 26.2 & 47.1 \\
\bottomrule
\end{tabular}

}
\end{minipage}

\subsection{PubHealthBench - results by topic area and audience}

\begin{figure}[ht]
  \centering%
  \begin{minipage}[t]{0.48\linewidth}
    \centering
    \includegraphics[width=\linewidth]{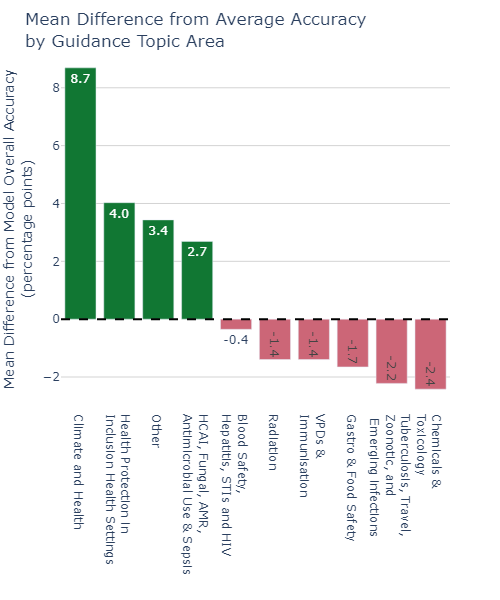}
  \end{minipage}%
  \begin{minipage}[t]{0.48\linewidth}
    \centering
    \includegraphics[width=\linewidth]{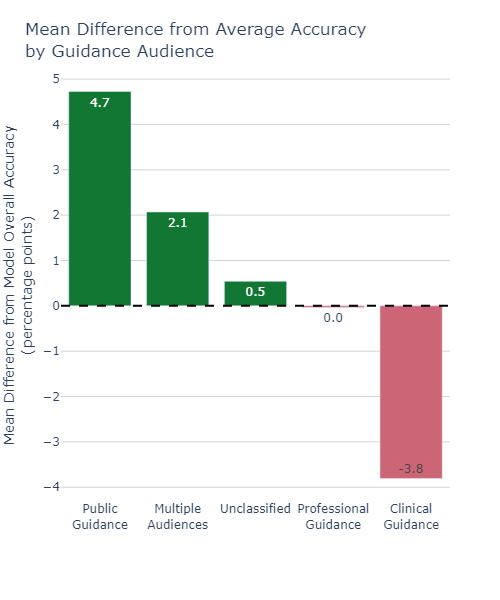}
  \end{minipage}%
  \caption{Average deviation from overall model performance on PubHealthBench-Full by guidance topic (left) and guidance audience (right).}\label{fig:structural_diff_combined}
\end{figure}

Breaking down PubHealthBench-Full results by topic area (Table~\ref{tab:results_full}), we find consistently higher performance across models on \textit{Climate and Health} and \textit{Health Protection in Inclusion Health Settings} guidance. While LLMs generally performed worse on \textit{Chemicals and Toxicology} (Figure~\ref{fig:structural_diff_combined}).

Looking at results by guidance audience (Table~\ref{tab:results_full_type}), we find LLMs have better knowledge of guidance intended for the general public, and worse knowledge of clinical guidance (Figure~\ref{fig:structural_diff_combined}). On public guidance the highest performing model (GPT-4.5) scores 96\%, close to the estimated upper bound. 

\subsection{PubHealthBench-FreeForm results}

The free form response setup is substantially more challenging for a few reasons: (1) it requires recalling the correct guidance information without any hints from MCQA options, (2) it introduces the possibility of LLMs hallucinating additional information that may be inconsistent with the source text, and (3) the correct answer cannot be inferred via elimination of the other answer options.  As a result, all models achieve substantially lower scores in the free form setting by up to 60 percentage points (ppts). The best performing model (o1) scores 74\% (Table~\ref{tab:judge_results}) and also sees the smallest decline from its MCQA performance at -17ppts (Table~\ref{tab:judge_results_diff}). We provide comparable results using three additional judge models in Appendix~\ref{supp:judge_setup}, finding a high level of agreement across judges.

Notably there is also significant variation across models with some smaller LLMs (e.g Phi-4-14B) showing over 45ppt declines from MCQA accuracy, while other similarly sized models (e.g Gemma-3-12B) see drops of comparable magnitude to SOTA proprietary LLMs. Also, importantly, as observed in the MCQA setting, LLMs consistently show more accurate knowledge of guidance intended for the general public compared to clinical or professional guidance.

\begin{table}[tbp]           
  \centering                 

  \begin{minipage}[t]{0.52\linewidth}
    \centering
    \captionof{table}{PubHealthBench-FreeForm accuracy by guidance audience. *LLM used to generate benchmark, **Judge LLM.}
      \vspace{1em} 
    \label{tab:judge_results}
    \resizebox{\linewidth}{!}{
\begin{tabular}{lcccccc}
\toprule
 & \makecell{Clinical\\Guidance} & \makecell{Multiple\\Audiences} & \makecell{Professional\\Guidance} & \makecell{Public\\Guidance} & \makecell{Unclassified} & \makecell{Total} \\
Model Name &  &  &  &  &  &  \\
\midrule
o1 & \textbf{71} & \textbf{81} & 70 & \textbf{86} & \textbf{82} & \textbf{74} \\
GPT-4.1 & 65 & 71 & \textbf{71} & 82 & 69 & 71 \\
o3-Mini & 65 & 78 & 69 & 78 & 74 & 70 \\
GPT-4o & 60 & 71 & 54 & 82 & 63 & 61 \\
GPT-4.5 & 59 & 71 & 54 & 77 & 59 & 59 \\
Claude-Sonnet-3.7 & 57 & 66 & 55 & 76 & 57 & 59 \\
Gemini-2.0-Flash & 56 & 64 & 53 & 77 & 61 & 58 \\
Gemma-3-27B & 50 & 61 & 53 & 73 & 52 & 55 \\
Gemini-Pro-1.5 & 46 & 58 & 51 & 63 & 61 & 53 \\
Gemma-3-12B & 50 & 53 & 48 & 74 & 55 & 52 \\
Claude-Haiku-3.5 & 51 & 61 & 40 & 68 & 47 & 48 \\
GPT-4o-Mini** & 37 & 54 & 40 & 68 & 41 & 43 \\
Llama-3.3-70B* & 35 & 53 & 38 & 60 & 40 & 41 \\
Mistral-3.1-24B & 36 & 39 & 36 & 64 & 40 & 40 \\
Phi-4-14B & 33 & 46 & 37 & 59 & 40 & 39 \\
Olmo-2-32B & 37 & 51 & 37 & 55 & 29 & 39 \\
Gemma-3-4B & 25 & 39 & 35 & 58 & 44 & 37 \\
Command-R-32B & 30 & 41 & 29 & 53 & 42 & 34 \\
Olmo-2-13B & 30 & 32 & 30 & 60 & 34 & 34 \\
Gemma-2-27B & 27 & 36 & 32 & 49 & 34 & 33 \\
Command-R-7B & 20 & 17 & 22 & 46 & 19 & 23 \\
Llama-3.1-8B & 16 & 22 & 15 & 38 & 18 & 19 \\
Phi-4-4B & 16 & 24 & 17 & 29 & 15 & 19 \\
Gemma-3-1B & 14 & 12 & 21 & 26 & 14 & 18 \\
\bottomrule
\end{tabular}

}
  \end{minipage}\hfill
  \begin{minipage}[t]{0.45\linewidth}
    \centering
    \captionof{table}{Difference in accuracy between MCQA and Free Form settings. *LLM used to generate benchmark, **Judge LLM.}
      \vspace{1em} 
    \label{tab:judge_results_diff}
    \resizebox{\linewidth}{!}{
\begin{tabular}{lcccccc}
\toprule
 & \makecell{PubHealthBench\\Reviewed} & \makecell{PubHealthBench\\FreeForm} & \makecell{MCQA - FreeForm\\Difference } \\
Model Name &  &  &   \\
\midrule
o1 & 91 & 74 & -17 \\
o3-Mini & 88 & 70 & -18 \\
GPT-4o & 80 & 60 & -19 \\
GPT-4.1 & 92 & 70 & -21 \\
Gemma-3-1B & 45 & 18 & -27 \\
Gemma-3-12B & 80 & 52 & -27 \\
Gemma-3-27B & 82 & 54 & -28 \\
Claude-Sonnet-3.7 & 87 & 58 & -29 \\
Gemini-2.0-Flash & 88 & 57 & -30 \\
Gemini-Pro-1.5 & 86 & 52 & -33 \\
GPT-4.5 & 92 & 59 & -33 \\
Claude-Haiku-3.5 & 83 & 48 & -35 \\
Gemma-3-4B & 73 & 36 & -36 \\
Olmo-2-32B & 78 & 38 & -39 \\
GPT-4o-Mini** & 83 & 43 & -40 \\
Olmo-2-13B & 75 & 33 & -41 \\
Mistral-3.1-24B & 84 & 39 & -45 \\
Llama-3.3-70B* & 87 & 40 & -46 \\
Command-R-32B & 80 & 34 & -46 \\
Phi-4-14B & 86 & 39 & -47 \\
Command-R-7B & 72 & 23 & -49 \\
Gemma-2-27B & 82 & 33 & -49 \\
Llama-3.1-8B & 81 & 18 & -62 \\
Phi-4-4B & 81 & 18 & -63 \\
\bottomrule
\end{tabular}

}
  \end{minipage}

\end{table}          

\section{Discussion}\label{sec:discussion}
Our results suggest that current SOTA LLMs, both proprietary and open-weight, in general have a very high level of knowledge across UK public health guidance.  This is particularly notable given 31\% of the MCQA questions were based on guidance documents that were at least partially updated within 2024 (see Appendix~\ref{supp:bench_stats}), after many of the LLMs' training data cut-off date. 

However, we find that performance is significantly degraded across models in the free form response setting. This is in part due to models including extraneous recommendations that do not form part of the source UK public health guidance, but main issues appear to be omitting or contradicting guidance information (see Appendix~\ref{sec:error_analysis}). Qualitative assessment of o1 responses suggests possible problematic outputs are often around the timing of interventions, we provide some examples of these in Appendix~\ref{supp:harm_exp}. 

Importantly for real-world use cases and deployments we see large disparities between the proprietary and large open-weight models compared with smaller open-weight LLMs (1-15bn parameters). On MCQA questions this gap is generally 10-20ppts but often grows to more than 35ppts in the free form setting. Therefore, there still appear to be significant risks around hallucinations relating to UK public health guidance when using smaller LLMs. 

From a public health perspective, it is also an important finding that LLMs consistently performed best on guidance intended for the general public. This audience is likely the highest risk set of users for querying chatbots to retrieve public health information. The fact LLMs are observed to have greater knowledge in this area implies the risks may be lower than the overall results would imply.

\section{Limitations}

Whilst we have attempted to include a broad range of topics and two LLM response formats (MCQA and free form), a limitation is that this still only represents a few of the ways LLMs could be used to retrieve public health information. Further work investigating different types of public health query is needed, for example, multi-turn interactions, queries including images, queries about topics that are related to public health but not directly incorporated into UK guidance, or allowing tool use.

We use English-language permissively-licensed online UK Government information, which may be incorporated into LLM training data. These results may not extrapolate to LLM performance on other countries' guidance or to other languages. While we hope as the first comprehensive benchmark in this area the approach and LLM performance has broader relevance, the specific results only directly relate to UK information. Further work extending the evaluations to other global guidance sources is needed, particularly for lower resource languages and other global health settings.

\section{Conclusion}

In this paper we use an automated pipeline to generate a new benchmark, PubHealthBench, to assess LLM knowledge of current UK Government public health guidance. We evaluate SOTA LLMs across three versions of the benchmark: full, verified, and free form. By testing LLMs across 10 guidance topic areas and 3 intended audiences in the MCQA and free form response setups, we provide an initial assessment of the performance and risks of using current LLMs in this domain. Our results suggest that the latest proprietary LLMs have a high degree of knowledge of UK public health guidance but challenges remain consistently matching the ground truth guidance in the free form setting. We hope this new benchmark will enable further research and evaluation of LLMs for public health, and reduce the risks within real-world deployments. 

\section{Ethics Statement}

All human annotators involved in quality assuring the benchmark were full time employees and paid at least the minimum wage. All data was sourced from the .gov.uk website with permissive licences (Open Government Licence 3.0) and also used in-line with the \href{https://www.gov.uk/help/reuse-govuk-content}{reuse GOV.UK content} terms of service. Large Language Models were used to help generate some diagrams and figures in this paper.

\section{ Reproducibility Statement}

To ensure reproducibility we provide the full benchmark dataset and example evaluation code as part of the supplementary materials and provide full details of the generation steps in Section \ref{sec:pipeline} and Appendix \ref{supp:bench_prompts}. To enable reproducibility of the free form LLM-as-a-Judge results we also include in the dataset the relevant chunks retrieved from the corpus and provided to the judge.



\bibliography{iclr2026/refs}

@article{chatgptUSMLE,
    doi = {10.1371/journal.pdig.0000198},
    author = {Kung, Tiffany H. AND Cheatham, Morgan AND Medenilla, Arielle AND Sillos, Czarina AND De Leon, Lorie AND Elepaño, Camille AND Madriaga, Maria AND Aggabao, Rimel AND Diaz-Candido, Giezel AND Maningo, James AND Tseng, Victor},
    journal = {PLOS Digital Health},
    publisher = {Public Library of Science},
    title = {Performance of ChatGPT on USMLE: Potential for AI-assisted medical education using large language models},
    year = {2023},
    month = {02},
    volume = {2},
    url = {https://doi.org/10.1371/journal.pdig.0000198},
    pages = {1-12},
    number = {2},

}

@article{singhal2023large_usmle,
  title={Large language models encode clinical knowledge},
  author={Singhal, Karan and Azizi, Shekoofeh and Tu, Tao and Mahdavi, S Sara and Wei, Jason and Chung, Hyung Won and Scales, Nathan and Tanwani, Ajay and Cole-Lewis, Heather and Pfohl, Stephen and others},
  journal={Nature},
  volume={620},
  number={7972},
  pages={172--180},
  year={2023},
  publisher={Nature Publishing Group}
}

@article{openai2023gpt4,
      title={GPT-4 Technical Report}, 
      author={OpenAI},
      year={2023},
      eprint={2303.08774},
      archivePrefix={arXiv},
      primaryClass={cs.CL},
journal={arXiv preprint arXiv:2303.08774},
}

@misc{hendrycks2021measuringmassivemultitasklanguage,
      title={Measuring Massive Multitask Language Understanding}, 
      author={Dan Hendrycks and Collin Burns and Steven Basart and Andy Zou and Mantas Mazeika and Dawn Song and Jacob Steinhardt},
      year={2021},
      eprint={2009.03300},
      archivePrefix={arXiv},
      primaryClass={cs.CY},
      url={https://arxiv.org/abs/2009.03300}, 
}

@misc{wang2024mmluprorobustchallengingmultitask,
      title={MMLU-Pro: A More Robust and Challenging Multi-Task Language Understanding Benchmark}, 
      author={Yubo Wang and Xueguang Ma and Ge Zhang and Yuansheng Ni and Abhranil Chandra and Shiguang Guo and Weiming Ren and Aaran Arulraj and Xuan He and Ziyan Jiang and Tianle Li and Max Ku and Kai Wang and Alex Zhuang and Rongqi Fan and Xiang Yue and Wenhu Chen},
      year={2024},
      eprint={2406.01574},
      archivePrefix={arXiv},
      primaryClass={cs.CL},
      url={https://arxiv.org/abs/2406.01574}, 
}

@misc{gema2024mmlu,
      title={Are We Done with MMLU?}, 
      author={Aryo Pradipta Gema and Joshua Ong Jun Leang and Giwon Hong and Alessio Devoto and Alberto Carlo Maria Mancino and Rohit Saxena and Xuanli He and Yu Zhao and Xiaotang Du and Mohammad Reza Ghasemi Madani and Claire Barale and Robert McHardy and Joshua Harris and Jean Kaddour and Emile van Krieken and Pasquale Minervini},
      year={2024},
      eprint={2406.04127},
      archivePrefix={arXiv},
      primaryClass={cs.CL},
      url={https://arxiv.org/abs/2406.04127}, 
}

@misc{rein2023gpqagraduatelevelgoogleproofqa,
      title={GPQA: A Graduate-Level Google-Proof Q\&A Benchmark}, 
      author={David Rein and Betty Li Hou and Asa Cooper Stickland and Jackson Petty and Richard Yuanzhe Pang and Julien Dirani and Julian Michael and Samuel R. Bowman},
      year={2023},
      eprint={2311.12022},
      archivePrefix={arXiv},
      primaryClass={cs.AI},
      url={https://arxiv.org/abs/2311.12022}, 
}

@misc{clark2018thinksolvedquestionanswering,
      title={Think you have Solved Question Answering? Try ARC, the AI2 Reasoning Challenge}, 
      author={Peter Clark and Isaac Cowhey and Oren Etzioni and Tushar Khot and Ashish Sabharwal and Carissa Schoenick and Oyvind Tafjord},
      year={2018},
      eprint={1803.05457},
      archivePrefix={arXiv},
      primaryClass={cs.AI},
      url={https://arxiv.org/abs/1803.05457}, 
}

@misc{pal2022medmcqalargescalemultisubject,
      title={MedMCQA : A Large-scale Multi-Subject Multi-Choice Dataset for Medical domain Question Answering}, 
      author={Ankit Pal and Logesh Kumar Umapathi and Malaikannan Sankarasubbu},
      year={2022},
      eprint={2203.14371},
      archivePrefix={arXiv},
      primaryClass={cs.CL},
      url={https://arxiv.org/abs/2203.14371}, 
}

@misc{jin2020diseasedoespatienthave,
      title={What Disease does this Patient Have? A Large-scale Open Domain Question Answering Dataset from Medical Exams}, 
      author={Di Jin and Eileen Pan and Nassim Oufattole and Wei-Hung Weng and Hanyi Fang and Peter Szolovits},
      year={2020},
      eprint={2009.13081},
      archivePrefix={arXiv},
      primaryClass={cs.CL},
      url={https://arxiv.org/abs/2009.13081}, 
}

@misc{bommarito2023gptknowledgeworkerzeroshot,
      title={GPT as Knowledge Worker: A Zero-Shot Evaluation of (AI)CPA Capabilities}, 
      author={Jillian Bommarito and Michael Bommarito and Daniel Martin Katz and Jessica Katz},
      year={2023},
      eprint={2301.04408},
      archivePrefix={arXiv},
      primaryClass={cs.CL},
      url={https://arxiv.org/abs/2301.04408}, 
}

@misc{guinet2024automatedevaluationretrievalaugmentedlanguage,
      title={Automated Evaluation of Retrieval-Augmented Language Models with Task-Specific Exam Generation}, 
      author={Gauthier Guinet and Behrooz Omidvar-Tehrani and Anoop Deoras and Laurent Callot},
      year={2024},
      eprint={2405.13622},
      archivePrefix={arXiv},
      primaryClass={cs.CL},
      url={https://arxiv.org/abs/2405.13622}, 
}

@misc{ghazaryan2024syndarinsynthesisingdatasetsautomated,
      title={SynDARin: Synthesising Datasets for Automated Reasoning in Low-Resource Languages}, 
      author={Gayane Ghazaryan and Erik Arakelyan and Pasquale Minervini and Isabelle Augenstein},
      year={2024},
      eprint={2406.14425},
      archivePrefix={arXiv},
      primaryClass={cs.CL},
      url={https://arxiv.org/abs/2406.14425}, 
}

@article{davies2024chatgpt,
  title={ChatGPT sits the DFPH exam: large language model performance and potential to support public health learning},
  author={Davies, Nathan P and Wilson, Robert and Winder, Madeleine S and Tunster, Simon J and McVicar, Kathryn and Thakrar, Shivan and Williams, Joe and Reid, Allan},
  journal={BMC Medical Education},
  volume={24},
  number={1},
  pages={57},
  year={2024},
  publisher={Springer}
}

@misc{harris2024evaluatinglargelanguagemodels,
      title={Evaluating Large Language Models for Public Health Classification and Extraction Tasks}, 
      author={Joshua Harris and Timothy Laurence and Leo Loman and Fan Grayson and Toby Nonnenmacher and Harry Long and Loes WalsGriffith and Amy Douglas and Holly Fountain and Stelios Georgiou and Jo Hardstaff and Kathryn Hopkins and Y-Ling Chi and Galena Kuyumdzhieva and Lesley Larkin and Samuel Collins and Hamish Mohammed and Thomas Finnie and Luke Hounsome and Steven Riley},
      year={2024},
      eprint={2405.14766},
      archivePrefix={arXiv},
      primaryClass={cs.CL},
      url={https://arxiv.org/abs/2405.14766}, 
}

@article{ayers2023evaluating,
  title={Evaluating artificial intelligence responses to public health questions},
  author={Ayers, John W and Zhu, Zechariah and Poliak, Adam and Leas, Eric C and Dredze, Mark and Hogarth, Michael and Smith, Davey M},
  journal={JAMA network open},
  volume={6},
  number={6},
  pages={e2317517--e2317517},
  year={2023},
  publisher={American Medical Association}
}

@misc{bbcaiassitantsarticle,
  author = "{BBC}",
  title = "{Representation of BBC News content in AI Assistants}",
  year = 2024,
  url = "https://www.bbc.co.uk/aboutthebbc/documents/bbc-research-into-ai-assistants.pdf",
  note = "[Online; accessed 13-Feb-2025]"
}

@misc{gemini_2_5,
	author = {Google},
	title = {Gemini 2.5: Our most intelligent AI model},
	year = {2025},
    note = {\href{https://blog.google/technology/google-deepmind/gemini-model-thinking-updates-march-2025/#gemini-2-5-thinking}{Gemini 2.5: Our most intelligent AI model}, Accessed: 03/04/2025}
}

@misc{o3_mini,
	author = {OpenAI},
	title = {OpenAI o3-mini},
	year = {2025},
    note = {\href{https://openai.com/index/openai-o3-mini/}{OpenAI o3-mini}, Accessed: 03/04/2025}
}

@misc{chatgpt_chatbot,
	author = {OpenAI},
	title = {Introducing ChatGPT},
	year = {2022},
    note = {\href{https://openai.com/index/chatgpt/}{Introducing ChatGPT}, Accessed: 08/04/2025}
}

@misc{claude_3_7,
	author = {Anthropic},
	title = {Claude 3.7 Sonnet},
	year = {2025},
    note = {\href{https://www.anthropic.com/claude/sonnet}{Claude 3.7 Sonnet}, Accessed: 03/04/2025}}

@misc{gpt_4_5,
	author = {OpenAI},
	title = {OpenAI GPT-4.5 System Card},
	year = {2025},
    note = {\href{https://cdn.openai.com/gpt-4-5-system-card-2272025.pdf}{GPT-4.5 System Card}, Accessed: 07/05/2025}}

@misc{o1_sys_card,
	author = {OpenAI},
	title = {OpenAI o1 System Card},
	year = {2024},
    note = {\href{https://cdn.openai.com/o1-system-card.pdf}{o1 System Card}, Accessed: 07/05/2025}}

@misc{ukhsa_ai_board,
    title = {UKHSA Advisory Board - Artificial Intelligence Discovery Exercise},
    author = {UKHSA},
    year = {2023},
    note = {\href{https://assets.publishing.service.gov.uk/media/6560be273d7741000d4201b0/AB-23-064_AI_Strategy_Development.pdf}{UKHSA Advisory Board: Artificial Intelligence Discovery Exercise}, Accessed: 03/04/2025}
}

@misc{gov_ai_op,
    title = {AI Opportunities Action Plan: government response},
    author = {Department for Science, Innovation and Technology},
    year = {2023},
    note = {\href{https://www.gov.uk/government/publications/ai-opportunities-action-plan-government-response/ai-opportunities-action-plan-government-response}{AI Opportunities Action Plan: government response}, Accessed: 03/04/2025}
}

@misc{grattafiori2024llama3herdmodels,
      title={The Llama 3 Herd of Models}, 
      author={Aaron Grattafiori and Abhimanyu Dubey and Abhinav Jauhri and Abhinav Pandey and Abhishek Kadian and Ahmad Al-Dahle and Aiesha Letman and Akhil Mathur and Alan Schelten and Alex Vaughan and Amy Yang and Angela Fan and Anirudh Goyal and Anthony Hartshorn and Aobo Yang and Archi Mitra and Archie Sravankumar and Artem Korenev and Arthur Hinsvark and Arun Rao and Aston Zhang and Aurelien Rodriguez and Austen Gregerson and Ava Spataru and Baptiste Roziere and Bethany Biron and Binh Tang and Bobbie Chern and Charlotte Caucheteux and Chaya Nayak and Chloe Bi and Chris Marra and Chris McConnell and Christian Keller and Christophe Touret and Chunyang Wu and Corinne Wong and Cristian Canton Ferrer and Cyrus Nikolaidis and Damien Allonsius and Daniel Song and Danielle Pintz and Danny Livshits and Danny Wyatt and David Esiobu and Dhruv Choudhary and Dhruv Mahajan and Diego Garcia-Olano and Diego Perino and Dieuwke Hupkes and Egor Lakomkin and Ehab AlBadawy and Elina Lobanova and Emily Dinan and Eric Michael Smith and Filip Radenovic and Francisco Guzmán and Frank Zhang and Gabriel Synnaeve and Gabrielle Lee and Georgia Lewis Anderson and Govind Thattai and Graeme Nail and Gregoire Mialon and Guan Pang and Guillem Cucurell and Hailey Nguyen and Hannah Korevaar and Hu Xu and Hugo Touvron and Iliyan Zarov and Imanol Arrieta Ibarra and Isabel Kloumann and Ishan Misra and Ivan Evtimov and Jack Zhang and Jade Copet and Jaewon Lee and Jan Geffert and Jana Vranes and Jason Park and Jay Mahadeokar and Jeet Shah and Jelmer van der Linde and Jennifer Billock and Jenny Hong and Jenya Lee and Jeremy Fu and Jianfeng Chi and Jianyu Huang and Jiawen Liu and Jie Wang and Jiecao Yu and Joanna Bitton and Joe Spisak and Jongsoo Park and Joseph Rocca and Joshua Johnstun and Joshua Saxe and Junteng Jia and Kalyan Vasuden Alwala and Karthik Prasad and Kartikeya Upasani and Kate Plawiak and Ke Li and Kenneth Heafield and Kevin Stone and Khalid El-Arini and Krithika Iyer and Kshitiz Malik and Kuenley Chiu and Kunal Bhalla and Kushal Lakhotia and Lauren Rantala-Yeary and Laurens van der Maaten and Lawrence Chen and Liang Tan and Liz Jenkins and Louis Martin and Lovish Madaan and Lubo Malo and Lukas Blecher and Lukas Landzaat and Luke de Oliveira and Madeline Muzzi and Mahesh Pasupuleti and Mannat Singh and Manohar Paluri and Marcin Kardas and Maria Tsimpoukelli and Mathew Oldham and Mathieu Rita and Maya Pavlova and Melanie Kambadur and Mike Lewis and Min Si and Mitesh Kumar Singh and Mona Hassan and Naman Goyal and Narjes Torabi and Nikolay Bashlykov and Nikolay Bogoychev and Niladri Chatterji and Ning Zhang and Olivier Duchenne and Onur Çelebi and Patrick Alrassy and Pengchuan Zhang and Pengwei Li and Petar Vasic and Peter Weng and Prajjwal Bhargava and Pratik Dubal and Praveen Krishnan and Punit Singh Koura and Puxin Xu and Qing He and Qingxiao Dong and Ragavan Srinivasan and Raj Ganapathy and Ramon Calderer and Ricardo Silveira Cabral and Robert Stojnic and Roberta Raileanu and Rohan Maheswari and Rohit Girdhar and Rohit Patel and Romain Sauvestre and Ronnie Polidoro and Roshan Sumbaly and Ross Taylor and Ruan Silva and Rui Hou and Rui Wang and Saghar Hosseini and Sahana Chennabasappa and Sanjay Singh and Sean Bell and Seohyun Sonia Kim and Sergey Edunov and Shaoliang Nie and Sharan Narang and Sharath Raparthy and Sheng Shen and Shengye Wan and Shruti Bhosale and Shun Zhang and Simon Vandenhende and Soumya Batra and Spencer Whitman and Sten Sootla and Stephane Collot and Suchin Gururangan and Sydney Borodinsky and Tamar Herman and Tara Fowler and Tarek Sheasha and Thomas Georgiou and Thomas Scialom and Tobias Speckbacher and Todor Mihaylov and Tong Xiao and Ujjwal Karn and Vedanuj Goswami and Vibhor Gupta and Vignesh Ramanathan and Viktor Kerkez and Vincent Gonguet and Virginie Do and Vish Vogeti and Vítor Albiero and Vladan Petrovic and Weiwei Chu and Wenhan Xiong and Wenyin Fu and Whitney Meers and Xavier Martinet and Xiaodong Wang and Xiaofang Wang and Xiaoqing Ellen Tan and Xide Xia and Xinfeng Xie and Xuchao Jia and Xuewei Wang and Yaelle Goldschlag and Yashesh Gaur and Yasmine Babaei and Yi Wen and Yiwen Song and Yuchen Zhang and Yue Li and Yuning Mao and Zacharie Delpierre Coudert and Zheng Yan and Zhengxing Chen and Zoe Papakipos and Aaditya Singh and Aayushi Srivastava and Abha Jain and Adam Kelsey and Adam Shajnfeld and Adithya Gangidi and Adolfo Victoria and Ahuva Goldstand and Ajay Menon and Ajay Sharma and Alex Boesenberg and Alexei Baevski and Allie Feinstein and Amanda Kallet and Amit Sangani and Amos Teo and Anam Yunus and Andrei Lupu and Andres Alvarado and Andrew Caples and Andrew Gu and Andrew Ho and Andrew Poulton and Andrew Ryan and Ankit Ramchandani and Annie Dong and Annie Franco and Anuj Goyal and Aparajita Saraf and Arkabandhu Chowdhury and Ashley Gabriel and Ashwin Bharambe and Assaf Eisenman and Azadeh Yazdan and Beau James and Ben Maurer and Benjamin Leonhardi and Bernie Huang and Beth Loyd and Beto De Paola and Bhargavi Paranjape and Bing Liu and Bo Wu and Boyu Ni and Braden Hancock and Bram Wasti and Brandon Spence and Brani Stojkovic and Brian Gamido and Britt Montalvo and Carl Parker and Carly Burton and Catalina Mejia and Ce Liu and Changhan Wang and Changkyu Kim and Chao Zhou and Chester Hu and Ching-Hsiang Chu and Chris Cai and Chris Tindal and Christoph Feichtenhofer and Cynthia Gao and Damon Civin and Dana Beaty and Daniel Kreymer and Daniel Li and David Adkins and David Xu and Davide Testuggine and Delia David and Devi Parikh and Diana Liskovich and Didem Foss and Dingkang Wang and Duc Le and Dustin Holland and Edward Dowling and Eissa Jamil and Elaine Montgomery and Eleonora Presani and Emily Hahn and Emily Wood and Eric-Tuan Le and Erik Brinkman and Esteban Arcaute and Evan Dunbar and Evan Smothers and Fei Sun and Felix Kreuk and Feng Tian and Filippos Kokkinos and Firat Ozgenel and Francesco Caggioni and Frank Kanayet and Frank Seide and Gabriela Medina Florez and Gabriella Schwarz and Gada Badeer and Georgia Swee and Gil Halpern and Grant Herman and Grigory Sizov and Guangyi and Zhang and Guna Lakshminarayanan and Hakan Inan and Hamid Shojanazeri and Han Zou and Hannah Wang and Hanwen Zha and Haroun Habeeb and Harrison Rudolph and Helen Suk and Henry Aspegren and Hunter Goldman and Hongyuan Zhan and Ibrahim Damlaj and Igor Molybog and Igor Tufanov and Ilias Leontiadis and Irina-Elena Veliche and Itai Gat and Jake Weissman and James Geboski and James Kohli and Janice Lam and Japhet Asher and Jean-Baptiste Gaya and Jeff Marcus and Jeff Tang and Jennifer Chan and Jenny Zhen and Jeremy Reizenstein and Jeremy Teboul and Jessica Zhong and Jian Jin and Jingyi Yang and Joe Cummings and Jon Carvill and Jon Shepard and Jonathan McPhie and Jonathan Torres and Josh Ginsburg and Junjie Wang and Kai Wu and Kam Hou U and Karan Saxena and Kartikay Khandelwal and Katayoun Zand and Kathy Matosich and Kaushik Veeraraghavan and Kelly Michelena and Keqian Li and Kiran Jagadeesh and Kun Huang and Kunal Chawla and Kyle Huang and Lailin Chen and Lakshya Garg and Lavender A and Leandro Silva and Lee Bell and Lei Zhang and Liangpeng Guo and Licheng Yu and Liron Moshkovich and Luca Wehrstedt and Madian Khabsa and Manav Avalani and Manish Bhatt and Martynas Mankus and Matan Hasson and Matthew Lennie and Matthias Reso and Maxim Groshev and Maxim Naumov and Maya Lathi and Meghan Keneally and Miao Liu and Michael L. Seltzer and Michal Valko and Michelle Restrepo and Mihir Patel and Mik Vyatskov and Mikayel Samvelyan and Mike Clark and Mike Macey and Mike Wang and Miquel Jubert Hermoso and Mo Metanat and Mohammad Rastegari and Munish Bansal and Nandhini Santhanam and Natascha Parks and Natasha White and Navyata Bawa and Nayan Singhal and Nick Egebo and Nicolas Usunier and Nikhil Mehta and Nikolay Pavlovich Laptev and Ning Dong and Norman Cheng and Oleg Chernoguz and Olivia Hart and Omkar Salpekar and Ozlem Kalinli and Parkin Kent and Parth Parekh and Paul Saab and Pavan Balaji and Pedro Rittner and Philip Bontrager and Pierre Roux and Piotr Dollar and Polina Zvyagina and Prashant Ratanchandani and Pritish Yuvraj and Qian Liang and Rachad Alao and Rachel Rodriguez and Rafi Ayub and Raghotham Murthy and Raghu Nayani and Rahul Mitra and Rangaprabhu Parthasarathy and Raymond Li and Rebekkah Hogan and Robin Battey and Rocky Wang and Russ Howes and Ruty Rinott and Sachin Mehta and Sachin Siby and Sai Jayesh Bondu and Samyak Datta and Sara Chugh and Sara Hunt and Sargun Dhillon and Sasha Sidorov and Satadru Pan and Saurabh Mahajan and Saurabh Verma and Seiji Yamamoto and Sharadh Ramaswamy and Shaun Lindsay and Shaun Lindsay and Sheng Feng and Shenghao Lin and Shengxin Cindy Zha and Shishir Patil and Shiva Shankar and Shuqiang Zhang and Shuqiang Zhang and Sinong Wang and Sneha Agarwal and Soji Sajuyigbe and Soumith Chintala and Stephanie Max and Stephen Chen and Steve Kehoe and Steve Satterfield and Sudarshan Govindaprasad and Sumit Gupta and Summer Deng and Sungmin Cho and Sunny Virk and Suraj Subramanian and Sy Choudhury and Sydney Goldman and Tal Remez and Tamar Glaser and Tamara Best and Thilo Koehler and Thomas Robinson and Tianhe Li and Tianjun Zhang and Tim Matthews and Timothy Chou and Tzook Shaked and Varun Vontimitta and Victoria Ajayi and Victoria Montanez and Vijai Mohan and Vinay Satish Kumar and Vishal Mangla and Vlad Ionescu and Vlad Poenaru and Vlad Tiberiu Mihailescu and Vladimir Ivanov and Wei Li and Wenchen Wang and Wenwen Jiang and Wes Bouaziz and Will Constable and Xiaocheng Tang and Xiaojian Wu and Xiaolan Wang and Xilun Wu and Xinbo Gao and Yaniv Kleinman and Yanjun Chen and Ye Hu and Ye Jia and Ye Qi and Yenda Li and Yilin Zhang and Ying Zhang and Yossi Adi and Youngjin Nam and Yu and Wang and Yu Zhao and Yuchen Hao and Yundi Qian and Yunlu Li and Yuzi He and Zach Rait and Zachary DeVito and Zef Rosnbrick and Zhaoduo Wen and Zhenyu Yang and Zhiwei Zhao and Zhiyu Ma},
      year={2024},
      eprint={2407.21783},
      archivePrefix={arXiv},
      primaryClass={cs.AI},
      url={https://arxiv.org/abs/2407.21783}, 
}

@misc{shashidhar2025yourbencheasycustomevaluation,
      title={YourBench: Easy Custom Evaluation Sets for Everyone}, 
      author={Sumuk Shashidhar and Clémentine Fourrier and Alina Lozovskia and Thomas Wolf and Gokhan Tur and Dilek Hakkani-Tür},
      year={2025},
      eprint={2504.01833},
      archivePrefix={arXiv},
      primaryClass={cs.CL},
      url={https://arxiv.org/abs/2504.01833}, 
}

@misc{pteam2025supergpqascalingllmevaluation,
      title={SuperGPQA: Scaling LLM Evaluation across 285 Graduate Disciplines}, 
      author={P Team and Xinrun Du and Yifan Yao and Kaijing Ma and Bingli Wang and Tianyu Zheng and King Zhu and Minghao Liu and Yiming Liang and Xiaolong Jin and Zhenlin Wei and Chujie Zheng and Kaixin Deng and Shawn Gavin and Shian Jia and Sichao Jiang and Yiyan Liao and Rui Li and Qinrui Li and Sirun Li and Yizhi Li and Yunwen Li and David Ma and Yuansheng Ni and Haoran Que and Qiyao Wang and Zhoufutu Wen and Siwei Wu and Tyshawn Hsing and Ming Xu and Zhenzhu Yang and Zekun Moore Wang and Junting Zhou and Yuelin Bai and Xingyuan Bu and Chenglin Cai and Liang Chen and Yifan Chen and Chengtuo Cheng and Tianhao Cheng and Keyi Ding and Siming Huang and Yun Huang and Yaoru Li and Yizhe Li and Zhaoqun Li and Tianhao Liang and Chengdong Lin and Hongquan Lin and Yinghao Ma and Tianyang Pang and Zhongyuan Peng and Zifan Peng and Qige Qi and Shi Qiu and Xingwei Qu and Shanghaoran Quan and Yizhou Tan and Zili Wang and Chenqing Wang and Hao Wang and Yiya Wang and Yubo Wang and Jiajun Xu and Kexin Yang and Ruibin Yuan and Yuanhao Yue and Tianyang Zhan and Chun Zhang and Jinyang Zhang and Xiyue Zhang and Xingjian Zhang and Yue Zhang and Yongchi Zhao and Xiangyu Zheng and Chenghua Zhong and Yang Gao and Zhoujun Li and Dayiheng Liu and Qian Liu and Tianyu Liu and Shiwen Ni and Junran Peng and Yujia Qin and Wenbo Su and Guoyin Wang and Shi Wang and Jian Yang and Min Yang and Meng Cao and Xiang Yue and Zhaoxiang Zhang and Wangchunshu Zhou and Jiaheng Liu and Qunshu Lin and Wenhao Huang and Ge Zhang},
      year={2025},
      eprint={2502.14739},
      archivePrefix={arXiv},
      primaryClass={cs.CL},
      url={https://arxiv.org/abs/2502.14739}, 
}

@article{Huang_2025,
   title={A Survey on Hallucination in Large Language Models: Principles, Taxonomy, Challenges, and Open Questions},
   volume={43},
   ISSN={1558-2868},
   url={http://dx.doi.org/10.1145/3703155},
   DOI={10.1145/3703155},
   number={2},
   journal={ACM Transactions on Information Systems},
   publisher={Association for Computing Machinery (ACM)},
   author={Huang, Lei and Yu, Weijiang and Ma, Weitao and Zhong, Weihong and Feng, Zhangyin and Wang, Haotian and Chen, Qianglong and Peng, Weihua and Feng, Xiaocheng and Qin, Bing and Liu, Ting},
   year={2025},
   month=jan, pages={1–55} }

@misc{minaee2025largelanguagemodelssurvey,
      title={Large Language Models: A Survey}, 
      author={Shervin Minaee and Tomas Mikolov and Narjes Nikzad and Meysam Chenaghlu and Richard Socher and Xavier Amatriain and Jianfeng Gao},
      year={2025},
      eprint={2402.06196},
      archivePrefix={arXiv},
      primaryClass={cs.CL},
      url={https://arxiv.org/abs/2402.06196}, 
}

@inproceedings{kwon2023efficient,
  title={Efficient Memory Management for Large Language Model Serving with PagedAttention},
  author={Woosuk Kwon and Zhuohan Li and Siyuan Zhuang and Ying Sheng and Lianmin Zheng and Cody Hao Yu and Joseph E. Gonzalez and Hao Zhang and Ion Stoica},
  booktitle={Proceedings of the ACM SIGOPS 29th Symposium on Operating Systems Principles},
  year={2023}
}

@misc{zheng2023judgingllmasajudgemtbenchchatbot,
      title={Judging LLM-as-a-Judge with MT-Bench and Chatbot Arena}, 
      author={Lianmin Zheng and Wei-Lin Chiang and Ying Sheng and Siyuan Zhuang and Zhanghao Wu and Yonghao Zhuang and Zi Lin and Zhuohan Li and Dacheng Li and Eric P. Xing and Hao Zhang and Joseph E. Gonzalez and Ion Stoica},
      year={2023},
      eprint={2306.05685},
      archivePrefix={arXiv},
      primaryClass={cs.CL},
      url={https://arxiv.org/abs/2306.05685}, 
}

@misc{gu2025surveyllmasajudge,
      title={A Survey on LLM-as-a-Judge}, 
      author={Jiawei Gu and Xuhui Jiang and Zhichao Shi and Hexiang Tan and Xuehao Zhai and Chengjin Xu and Wei Li and Yinghan Shen and Shengjie Ma and Honghao Liu and Saizhuo Wang and Kun Zhang and Yuanzhuo Wang and Wen Gao and Lionel Ni and Jian Guo},
      year={2025},
      eprint={2411.15594},
      archivePrefix={arXiv},
      primaryClass={cs.CL},
      url={https://arxiv.org/abs/2411.15594}, 
}

@article{gemma_2025,
    title={Gemma 3},
    url={https://goo.gle/Gemma3Report},
    publisher={Kaggle},
    author={Gemma Team},
    year={2025}
}

@misc{abdin2024phi4technicalreport,
      title={Phi-4 Technical Report}, 
      author={Marah Abdin and Jyoti Aneja and Harkirat Behl and Sébastien Bubeck and Ronen Eldan and Suriya Gunasekar and Michael Harrison and Russell J. Hewett and Mojan Javaheripi and Piero Kauffmann and James R. Lee and Yin Tat Lee and Yuanzhi Li and Weishung Liu and Caio C. T. Mendes and Anh Nguyen and Eric Price and Gustavo de Rosa and Olli Saarikivi and Adil Salim and Shital Shah and Xin Wang and Rachel Ward and Yue Wu and Dingli Yu and Cyril Zhang and Yi Zhang},
      year={2024},
      eprint={2412.08905},
      archivePrefix={arXiv},
      primaryClass={cs.CL},
      url={https://arxiv.org/abs/2412.08905}, 
}

@misc{olmo20252olmo2furious,
      title={2 OLMo 2 Furious}, 
      author={Team OLMo and Pete Walsh and Luca Soldaini and Dirk Groeneveld and Kyle Lo and Shane Arora and Akshita Bhagia and Yuling Gu and Shengyi Huang and Matt Jordan and Nathan Lambert and Dustin Schwenk and Oyvind Tafjord and Taira Anderson and David Atkinson and Faeze Brahman and Christopher Clark and Pradeep Dasigi and Nouha Dziri and Michal Guerquin and Hamish Ivison and Pang Wei Koh and Jiacheng Liu and Saumya Malik and William Merrill and Lester James V. Miranda and Jacob Morrison and Tyler Murray and Crystal Nam and Valentina Pyatkin and Aman Rangapur and Michael Schmitz and Sam Skjonsberg and David Wadden and Christopher Wilhelm and Michael Wilson and Luke Zettlemoyer and Ali Farhadi and Noah A. Smith and Hannaneh Hajishirzi},
      year={2025},
      eprint={2501.00656},
      archivePrefix={arXiv},
      primaryClass={cs.CL},
      url={https://arxiv.org/abs/2501.00656}, 
}

@misc{asiedu2025contextualevaluationlargelanguage,
      title={Contextual Evaluation of Large Language Models for Classifying Tropical and Infectious Diseases}, 
      author={Mercy Asiedu and Nenad Tomasev and Chintan Ghate and Tiya Tiyasirichokchai and Awa Dieng and Oluwatosin Akande and Geoffrey Siwo and Steve Adudans and Sylvanus Aitkins and Odianosen Ehiakhamen and Eric Ndombi and Katherine Heller},
      year={2025},
      eprint={2409.09201},
      archivePrefix={arXiv},
      primaryClass={cs.CL},
      url={https://arxiv.org/abs/2409.09201}, 
}

@misc{bowyer2025positiondontuseclt,
      title={Position: Don't use the CLT in LLM evals with fewer than a few hundred datapoints}, 
      author={Sam Bowyer and Laurence Aitchison and Desi R. Ivanova},
      year={2025},
      eprint={2503.01747},
      archivePrefix={arXiv},
      primaryClass={cs.AI},
      url={https://arxiv.org/abs/2503.01747}, 
}

@misc{arora2025healthbenchevaluatinglargelanguage,
      title={HealthBench: Evaluating Large Language Models Towards Improved Human Health}, 
      author={Rahul K. Arora and Jason Wei and Rebecca Soskin Hicks and Preston Bowman and Joaquin Quiñonero-Candela and Foivos Tsimpourlas and Michael Sharman and Meghan Shah and Andrea Vallone and Alex Beutel and Johannes Heidecke and Karan Singhal},
      year={2025},
      eprint={2505.08775},
      archivePrefix={arXiv},
      primaryClass={cs.CL},
      url={https://arxiv.org/abs/2505.08775}, 
}
\bibliographystyle{iclr2026_conference}

\appendix
\section{Appendix}
\normalsize
\subsection{Automated MCQA error detection}\label{supp:auto_error}

Our approach to automated filtering of candidate questions leverages the fact that synthetic MCQA generation allows us to create an essentially arbitrary number of potential questions but with the risk of material question errors. An advantage of this is that we can utilise an LLM classier with high recall (accurately identifying incorrect questions) even if this comes at the expense of low precision (large numbers of valid questions being rejected) to filter our question set and ensure fewer material errors in the final benchmark.

To assess whether we can use LLMs for this error classification task on our generated MCQA questions we first manually annotated an evaluation dataset of LLM generated samples. We use samples from the earliest version of the pipeline that contained a much higher question error rate to ensure we had enough positive and negative samples. We build on the categorisation approach developed by \citet{gema2024mmlu} to annotate each question with one of 5 categories spanning the common errors that can be found in MCQA questions (the reviewer should allocate the first label that applies):

\begin{enumerate}
    \item \textbf{Valid Question and Options} \\
    The question contains all the required context to be able to answer the question and it is clear what information the question is seeking.

    \item \textbf{Ambiguous Question} \\
    The question cannot be answered standalone for some reason, including:
    \begin{enumerate}
        \item It is missing context necessary to answer the question (e.g., the disease the question refers to, the population the question refers to, etc.).
        \item It is poorly formed or does not make sense (e.g., asking about a made-up country).
        \item It contains material grammatical or typographical errors.
    \end{enumerate}

    \item \textbf{Ambiguous Options} \\
    The question is valid but it is not possible to determine if there is a correct option for some reason, including:
    \begin{enumerate}
        \item The options do not make sense given the question.
        \item The options do not contain enough context to determine whether any are correct.
    \end{enumerate}

    \item \textbf{Incorrect Answer} \\
    The question is valid and the options are clear but the proposed correct answer (a.) is not a valid correct answer to the question, independent of the other options provided.

    \item \textbf{Multiple Correct Answers} \\
    The question is valid, the options are clear, and the proposed correct answer (a.) is true, but there are also other options that would be \textbf{equally} valid.
\end{enumerate}

Two human experts then reviewed a random sample of 150 generated questions. We found the level of ambiguity and distractor option differentiation that was permissible for a valid question a challenging task even for expert human annotators. On the 150 questions inter-annotator agreement was 81\% with a Cohen's Kappa of 0.39. To create a final validation set of annotations all disagreements were reviewed and resolved by a panel of the two reviewers and a 3rd expert.

\begin{figure}[H]
    \centering
    \includegraphics[width=0.5\linewidth]{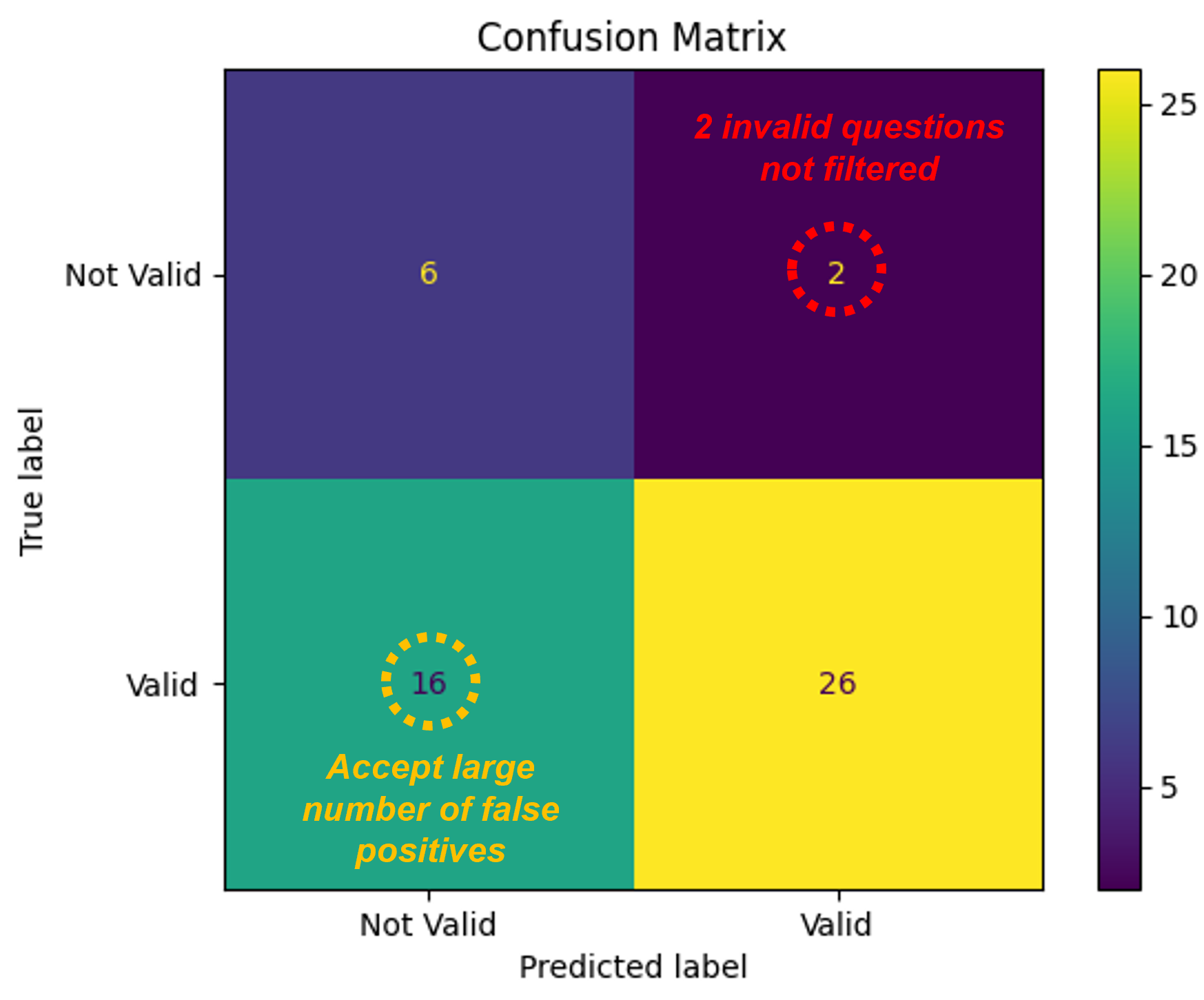} 
    \caption{Confusion matrix for LLM MCQA error detection.} \label{fig:error_detection}
\end{figure}

For the binary classification task of valid vs invalid, on the test set (of low quality questions) shown in Figure \ref{fig:error_detection} using Llama-3-70bn, this approach would in theory of reduced the benchmark error rate from approximately 16\% to approximately 8\%. However, we note that the limited test sample means further evaluation of approaches is needed. 

In our final pipeline, this stage was less critical as improving both the prompting approach and the LLM used in the generation process (Llama-3 to Llama-3.3) substantially reduced the rate of invalid MCQA samples generated by the pipeline. This is illustrated by the fact only approximately 8\% of generated MCQA samples were classified as invalid by the LLM error detection step in the final pipeline, compared with 44\% in the validation test set which was sampled from an early version. 
\newpage
\subsection{Additional benchmark statistics}\label{supp:bench_stats}

\begin{figure}[H]
    \centering
    \includegraphics[width=\linewidth]{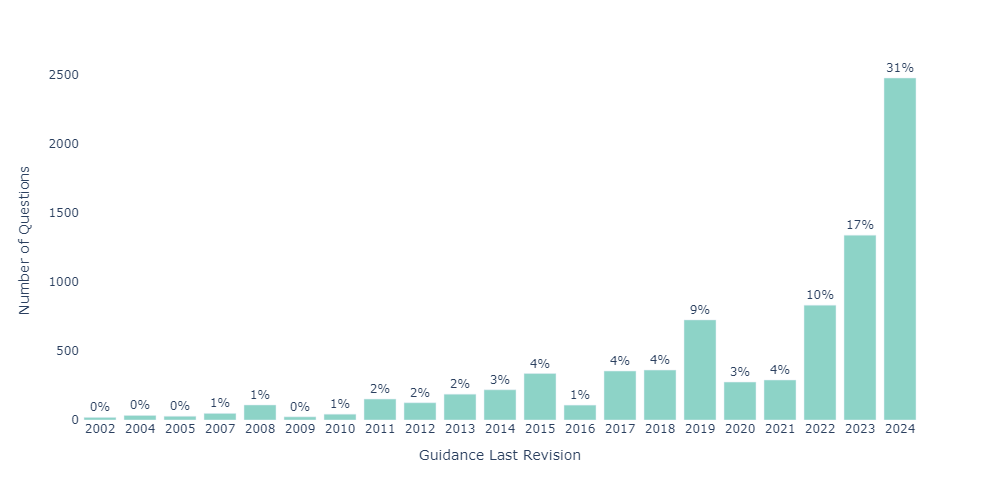} 
    \caption{PubHealthBench-Full MCQA by source guidance document last revision date. Note - the last revision to a document may only entail minor changes from the original text.} \label{fig:date_props}
\end{figure}

\begin{figure}[ht]
  \centering
  \begin{minipage}[t]{0.5\linewidth}
    \centering
    \includegraphics[width=\linewidth]{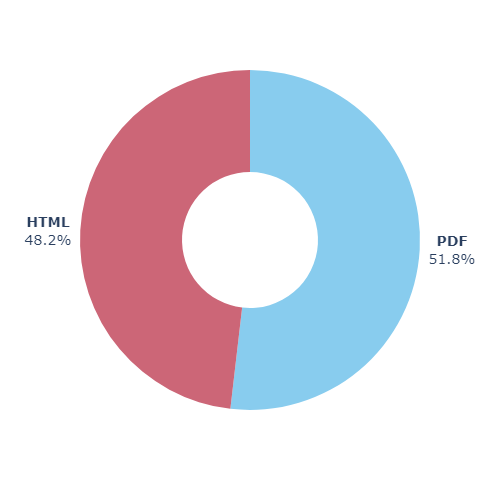}
  \end{minipage}\hfill
  \begin{minipage}[t]{0.5\linewidth}
    \centering
    \includegraphics[width=\linewidth]{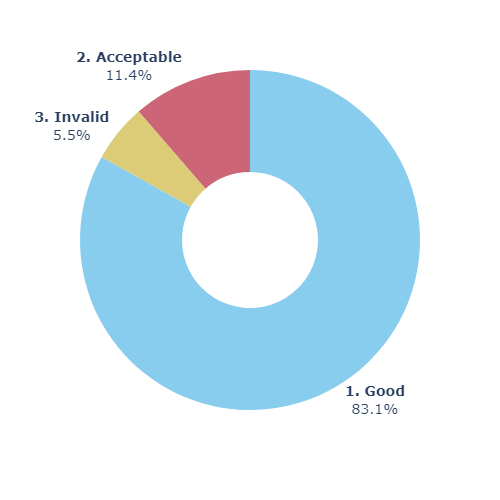}
  \end{minipage}
  \caption{(left) PubHealthBench-Full samples by source text document type, (right) MCQA quality human annotations - "Good" and "Acceptable" categories are treated as valid question samples.}
  \label{fig:human_annotation_fig}
\end{figure}

\subsection{MCQA generation prompts}\label{supp:generation_prompts}

\begin{table}[H]
    \caption{MCQA generation system prompt (see Table \ref{table:gen_body_prompt} for user prompt).}
    \label{table:gen_sys_prompt}
    \vspace{1em} 
    \centering
    \small
        \resizebox{0.96\linewidth}{!}{%
    \begin{tabular}{p{0.9\linewidth}}
\toprule
\textbf{System Prompt} \\ \midrule
You are a top public health expert, creating a multiple choice test to assess individuals' knowledge of UK Public Health Guidance - you should select questions relating to important parts of the information in the guidance text provided that have material public health implications. The questions should not relate to minor details such as phone numbers or other inconsequential information.
\\
\\
To make the questions in the test you are going to be given a piece of current UKHSA guidance, within which there will be a highlighted passage \textless\textless START OF SOURCE TEXT THE QUESTION SHOULD RELATE TO \textgreater\textgreater~[PASSAGE] \textless\textless END OF SOURCE TEXT THE QUESTION SHOULD RELATE TO \textgreater\textgreater. You will then do five things:
\\
\\
1. You will provide a list of bullet points of the key parts of highlighted passage in the guidance text provided that could be potentially important public health information and explain why they could be important. These bullet points should entirely come from the highlighted passage in the guidance text provided, you should not include any context or information not contained within the highlighted passage in the guidance text. 

2. You will then make a list of bullet points of the relevant pieces of contextual information contained in the highlighted passage of the source text that are necessary to include in future questions - THIS INFORMATION MUST BE INCLUDED IN THE QUESTION ITSELF UNLESS IT IS IRRELEVANT.

3. You will generate a question based on the information in the highlighted passage of the text provided (no other information should be needed to give the correct answer). The public health expert being asked the question will not have the source text and so YOU MUST ENSURE that all relevant context required is included in the question (e.g the names of the dieseases the question relates to, the sub-populations being disucssed, the geographical area the guidance relates to, the time period if specified, etc.).

4. You will then generate 7 multiple choice answers (a. to g.). The correct answer should always be the first answer (a.) you provide. The other answers should be incorrect but very plausible and challenging even for a public health expert to really test whether they actually know the fact. To construct the distractors you should draw on information in both the highlighted passage and the wider guidance provided.

5. Repeat steps 3., and 4., in order to generate a total of 2 questions and answers.
\\
\\
You should provide your final questions and options in the following JSON format:
\\
\\
\{\{\\
"0": \{\{"question": [INSERT QUESTION],\\
"a": [INSERT CORRECT ANSWER],\\
"b": [INSERT INCORRECT ANSWER],\\
"c": [INSERT INCORRECT ANSWER],\\
"d": [INSERT INCORRECT ANSWER],\\
"e": [INSERT INCORRECT ANSWER],\\
"f": [INSERT INCORRECT ANSWER],\\
"g": [INSERT INCORRECT ANSWER]\\
\}\},\\
"1" : \{\{"question": [INSERT QUESTION],\\
"a": [INSERT CORRECT ANSWER],\\
"b": [INSERT INCORRECT ANSWER],\\
"c": [INSERT INCORRECT ANSWER],\\
"d": [INSERT INCORRECT ANSWER],\\
"e": [INSERT INCORRECT ANSWER],\\
"f": [INSERT INCORRECT ANSWER],\\
"g": [INSERT INCORRECT ANSWER]\\
\}\}\\
\}\}\\
\bottomrule
\end{tabular}

    }
    \vspace{1em} 
\end{table}

\begin{table}[H]
    \caption{MCQA generation user prompt (see Table \ref{table:gen_sys_prompt} for system prompt).}
\label{table:gen_body_prompt}
    \vspace{1em} 
    \centering
    \small
        \resizebox{0.98\linewidth}{!}{%
    \begin{tabular}{p{0.9\linewidth}}
\toprule
\textbf{Prompt Content} \\ \midrule
IMPORTANT NOTES YOU MUST FOLLOW:\\
1. No real phone numbers or urls from the text should be included.\\
2. In both the question and the answer options you generate you should NOT mention anything relating to the structure of OTHER parts of the source text not included below, for example you should NOT mention things like: other sections of the text (e.g "refer to section 10"), question numbers (e.g "if the patient has answered yes to question 1"), further reading (e.g "see appendix for more information"), etc..\\
3. Be very careful not to include correct answers in the distractor options b. to g. (or distractors that are potentially correct based on your wider knowledge).\\
4. You must only generate a question specifically about the passage marked using: \textless\textless~ START OF SOURCE TEXT THE QUESTION SHOULD RELATE TO \textgreater\textgreater~ [PASSAGE] \textless\textless~ END OF SOURCE TEXT THE QUESTION SHOULD RELATE TO \textgreater\textgreater. You should only use the information outside of this passage for context. \\
5. Make sure you include all the relevant context in both the questions you generate, these questions will be separated and so should be totally independent.\\

Here is an example you should use as a template for the structure and style:\\

================================\\
\{one shot CoT example\}\\
================================\\

Now please follow the instructions above and generate the question and answer options for this piece of UKHSA guidance.\\

Guidance text: \\
\{guidance text\}\\
\\
Answer (Provide the bulleted list, contextual information, then the final JSON):\\
\\ \bottomrule
\end{tabular}
    }
    \vspace{1em} 
\end{table}

\subsection{Human manual annotation}\label{sec:supp_human_annotation}
In this work we perform two sets of human annotation, the first to estimate the rate of invalid or ambiguous questions, and the second to set a human baseline. In both cases, all annotators were full time employees within the organisation with relevant data science or public health experience. 

\subsubsection{Estimating the benchmark error rate}

Assessing MCQA sample validity is a challenging annotation task. For a question to be valid, judging the level of context (e.g subpopulation, geography, time period etc.) that is required for the question to be answerable can involve a significant degree of subjectivity. For the options to be valid, the boundary between when a challenging distractor option crosses over into being a potentially equally valid answer to the specified correct answer, rendering the question ambiguous, is also subjective. Therefore, we followed a two round annotation process.

In the first round, pairs of human experts were assigned a total of 150 MCQA samples for double review. Any discrepancies in annotations between reviewers were then assessed by all reviewers and the correct final annotation agreed. This enabled us to identify and rectify inconsistencies in the application of the annotation protocol. In the second round, the remaining 650 MCQA samples were then annotated by single reviewers who participated in the first round. 

We provide the Wilson score 95\% confidence interval for a binomial proportion. The full instructions provided to reviewers are shown in the box below:

\fbox{%
  \begin{minipage}{0.99\textwidth}

\textbf{Protocol for Manual Review of Exam Questions}\newline

This protocol provides structured criteria for assessing the validity of guidance LLM benchmark questions and answer options into three categories: \textbf{Good}, \textbf{Acceptable}, and \textbf{Incorrect}.

\vspace{1em}

\textbf{1. Good Questions}

\textit{Criteria:}
\begin{enumerate}
    \item \textbf{Question Valid}: The question is answerable and clearly aligned with the guidance document.
    \item \textbf{Best Answer Clearly Identifiable}: The correct answer is evidently the ‘best’ choice compared to the other options and is similar to a hypothetical ‘gold standard’ answer.
    \item \textbf{Other Options Incorrect but Not Trivially Wrong}: At least some of the other incorrect options are plausible, not wrong by definition, or so obviously wrong that an uninformed member of the public could say that is not something guidance would ever say.
    \item \textbf{Informative Value}: The LLM’s performance on this question adds meaningful information about the LLM’s knowledge of the guidance.
\end{enumerate}

\vspace{1em}

\textbf{2. Acceptable Questions}

\textit{Definition:} Questions in this category are valid but have limitations that reduce their overall quality or informativeness. These questions are still useful but might not be as robust as "Good" questions.

\textit{Criteria:}
\begin{enumerate}
    \item \textbf{Question Valid but Some Ambiguity}: The question can be understood and is answerable, but it may be missing some context, contain uncommon acronyms, or may assume a high degree of knowledge.
    \item \textbf{Gaps in Correct Answer but Still the Best Option}: The correct answer is the best choice from the options provided, but may not be the perfect gold standard answer, may lack some detail or nuance from the guidance, and may be poorly phrased.
    \item \textbf{Other Options Incorrect but Potentially Trivially Wrong}: All incorrect options are worse options than the correct answer but they may be incorrect trivially, making the correct answer easy to guess.
    \item \textbf{Some Relevance but Lower Informative Value}: The LLM’s performance on this question has some relevance (even if minor) but may test less critical aspects of the guidance, or the question may cover overlapping points with others in the dataset. Very easy questions would also be acceptable.
    \item \textbf{Not Directly Aligned with the Chunk Provided}: In some cases, the question may relate to text found either side of the intended chunk of guidance in the document. This is acceptable so long as the question still meets the criteria above.
\end{enumerate}

\vspace{1em}

\textbf{3. Invalid Questions}

\textit{Definition:} Questions in this category cannot be answered due to material errors and are unsuitable for use in the evaluation.

\textit{Criteria:}
\begin{enumerate}
    \item \textbf{Invalidity}: The question is not answerable due to ambiguity, misinterpretation of guidance, or grammatical errors.
    \item \textbf{Misleading Answer Options}: The correct answer option provided is either:
    \begin{enumerate}
        \item Not a valid answer to the question, or
        \item Clearly a less accurate answer than one or more of the distractor options.
    \end{enumerate}
    \item \textbf{Errors in Construction}: The question has structural or logical flaws that render it unusable.
\end{enumerate}

  \end{minipage}%
}

\subsubsection{Human baseline}

To set the human baseline 600 questions were randomly sampled from the manually reviewed subset of 800 questions. To make the human baseline comparable to the full benchmark scores these samples \textit{included} MCQA examples that human annotators had classified as invalid.

The 600 questions were then randomly allocated among 5 human test takers who had not manually reviewed the questions previously. Results were then aggregated to provide the human baseline. Test takers were provided with the following instructions:\newline

\fbox{%
  \begin{minipage}{0.99\textwidth}
\textbf{Human Test Taking Instructions}\newline\newline
These multiple choice questions are designed to assess LLM knowledge of UKHSA public health guidance. Understanding the performance of humans on these tests is useful for two reasons:

\begin{enumerate}
    \item To provide a baseline for the overall difficulty of the questions.
    \item To understand whether a human asking a chatbot (an LLM) a public health related question is higher or lower risk than a human using previous tools (e.g., Google) to try to find the same information themselves.
\end{enumerate}

To answer the questions, please follow these instructions:\newline

\textbf{Question Format}
\begin{enumerate}
    \item \textbf{Standard Multiple Choice} – Every question has one correct answer option in the provided options list.
    \item \textbf{Select Best Answer} – There may be multiple options that are technically valid information; you should select the answer that is the best reflection of UKHSA guidance.
    \item \textbf{Don't Know} – For any questions where you aren't confident in the answer, you should provide your best guess.
\end{enumerate}

\textbf{Test Conditions}
\begin{enumerate}
    \item \textbf{Timing} – Whilst there is no time constraint on completing the test, we recommend spending no more than 2 minutes on average per question.
    \item \textbf{Open Book} – Feel free to use your own knowledge and any publicly available online guidance resources to answer the questions.
    \item \textbf{No LLM Tools} – Please do not use any LLM-based tools or chatbots to search for guidance or question answers.
\end{enumerate}

  \end{minipage}%
}

\subsection{Experimental setup and compute resources}\label{sec:supp_setup}

For all experiments and models where it can be specified we use greedy decoding (temperature 0). For all OpenAI reasoning models we use "reasoning\_effort=low" for cost reasons and due to the knowledge based nature of the task.

Proprietary models are accessed via provider APIs, open-weight models are hosted internally on HPC resources using vLLM~\citep{kwon2023efficient}, with up to 2 x A100 80GB used per model. Additional compute (also using 2x A100 80GB) was used for benchmark generation pipeline.

We used a budget of \$400 for the proprietary model evaluations, including the LLM judge.

\subsection{PubHealthBench prompts}\label{supp:bench_prompts}

\begin{table}[H]
    \caption{MCQA zero-shot evaluation prompt.}
    \vspace{1em} 
    \centering
    \small
        \resizebox{0.98\linewidth}{!}{%
    \begin{tabular}{p{0.9\linewidth}}
\toprule
\textbf{System Prompt} \\ \midrule
You are an expert working for a Public Health agency.\\
\\
\midrule
\textbf{Prompt Content} \\ \midrule
The following are multiple choice questions (with answers) about UK Government public health guidance.\\ \\ Question: This question relates to UK Health Security Agency (UKHSA) guidance that could be found on the gov.uk website as of 08/01/2025.\\ \\ \{question\}\\ \\ Options:\\ \{answer\_options\}\\ \\ Provide the letter (A, B, C, D, E, F, or G) of the correct answer. You should state "The answer is (X)", where the X contained in the brackets is the correct letter choice, make sure you include the brackets () around your final answer in your response. DO NOT provide any other information or text in your response.\\ \\ Answer: 
\\ \bottomrule
\end{tabular}
    }
    \vspace{1em} 
\end{table}

\begin{table}[H]
    \caption{Free form evaluation prompt.}
    \vspace{1em} 
    \centering
    \small
        \resizebox{0.98\linewidth}{!}{%
    \begin{tabular}{p{0.9\linewidth}}
\toprule
\textbf{System Prompt} \\ \midrule
You are an expert working for a Public Health agency.\\
\\
\midrule
\textbf{Prompt Content} \\ \midrule
The following is a question about UK Government public health guidance.\\ \\ Question: This question relates to UK Health Security Agency (UKHSA) guidance. \\ \\
\{question\}\\ \\ Please answer the question to the best of your knowledge.\\ \\ Answer: 
\\ \bottomrule
\end{tabular}
    }
    \vspace{1em} 
\end{table}

\subsection{Free form evaluation - LLM as a Judge setup}\label{supp:judge_setup}

To provide an accurate but cost effective LLM judge to assess unstructured free form LLM responses, we frame it as a binary classification task for whether the free form responses align with the correct answer and source text.

We provide the judge with the guidance source text that was used to generate the question. This ensures the judge always has access to the ground truth information. To allow the judge to correctly evaluate answers that may include additional information that relates to other aspects of UK public health guidance, we provide 5 additional chunks of relevant guidance in the prompt. These 5 additional chunks for each question are retrieved from the full corpus using a hybrid retrieval system that ranked relevance by combining the cosine similarities of text embedding vectors (OpenAI's text-3-embedding-large model) and TF-IDF vectors, between each chunk and a given question.

Finally, to guide the judge to the most important information to include, we also provide the MCQA correct answer option. The full LLM judge CoT prompt combining these components is shown in Table \ref{tab:llm_judge_prompt}. 

 To create a judge evaluation dataset and assess the performance of the LLM judge, we utilise the already generated correct and incorrect MCQA answer options. We insert the known correct or incorrect answer option into a variety of free form chat response templates designed to simulate the structure of LLM free form responses. We assess the ability of the judge to distinguish whether the response is valid using the provided source text. 

In this paper we use gpt-4o-mini-2024-07-18 as the main judge with greedy decoding, which achieved over 99\% accuracy on the judge evaluation set (10,517 samples). Further details of this LLM judge evaluation approach will be published in upcoming work.  

\begin{table}[htbp]
    \caption{Judge prompt.}
    \vspace{1em} 
    \label{tab:llm_judge_prompt}
    \centering
    \small
        \resizebox{0.98\linewidth}{!}{%
    \begin{tabular}{p{0.9\linewidth}}
\toprule
\textbf{System Prompt} \\ \midrule
You are an expert in UK public health. You are going to evaluate whether a given answer to a public health guidance question is correct.\\
\\
\midrule
\textbf{Prompt Content} \\ \midrule
You are tasked with evaluating whether a given answer is correct based on the ground truth answer and provided context. Carefully analyse the ground truth answer and context and determine whether the given answer correctly answers the question and aligns with the information given.\\ ---\\===========\\ Question:\\ ===========\\ \{question\}\\ ---\\ \\ ===========\\ Context:\\ ===========\\ \{context\}\\ ---\\ \\ ===========\\ Ground Truth Answer:\\ ===========\\ \{ground truth answer\}\\ ---\\ \\ ===========\\ Given Answer:\\ ===========\\ \{given answer\}\\ ---\\ \\ For the given answer to be correct it must align with the ground truth without omitting any key details **and** any additional detail in the given answer must be seen in the provided context. Determine, with reasoning, whether the given answer is correct based on the ground truth answer and context. Give your response in the following json format:\\ \\ \{\{"reasoning": Why is the answer correct/incorrect, "predicted\_correct": true or false\}\}
\\ \bottomrule
\end{tabular}
    }
    \vspace{1em} 
\end{table}

\subsubsection{Judge model comparisons}

\begin{figure}[H]
    \centering
    \includegraphics[width=\linewidth]{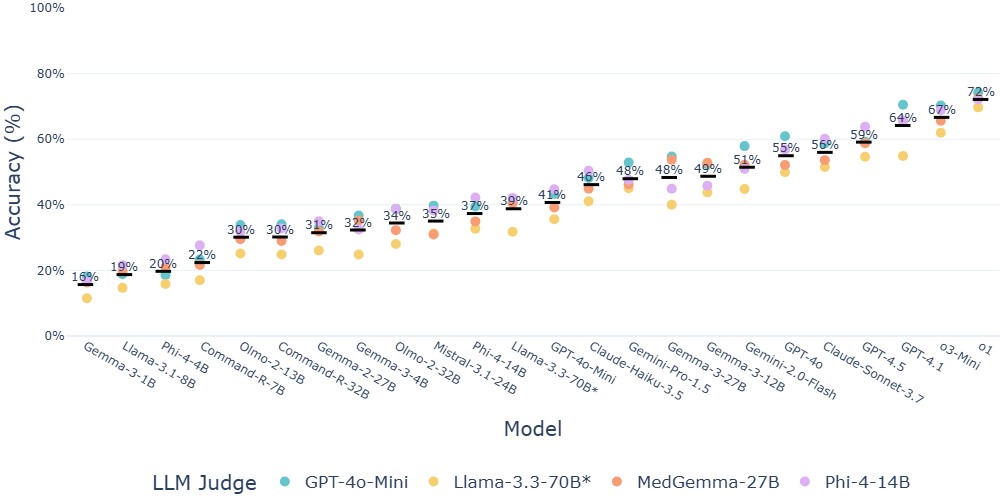} 
    \caption{Comparison of PubHealthBench-FreeForm performance using different LLM judges. *LLM used to generate benchmark.} \label{fig:judge_comparison}
\end{figure}

To compare judge models and assess the sensitivity of LLM performance to the judge model used we also re-run the free form evaluation using three additional judge models with range of model sizes. In addition to GPT-4o-Mini, we use Llama-3.3-70bn, MedGemma-27bn, and Phi-4-14bn. Overall we find a high degree of agreement across the four judge models, with a Fleiss's kappa of 0.64, see Figure \ref{fig:judge_comparison} for full results.

\subsubsection{Error analysis}\label{sec:error_analysis}

\begin{figure}[H]
    \centering
    \includegraphics[width=\linewidth]{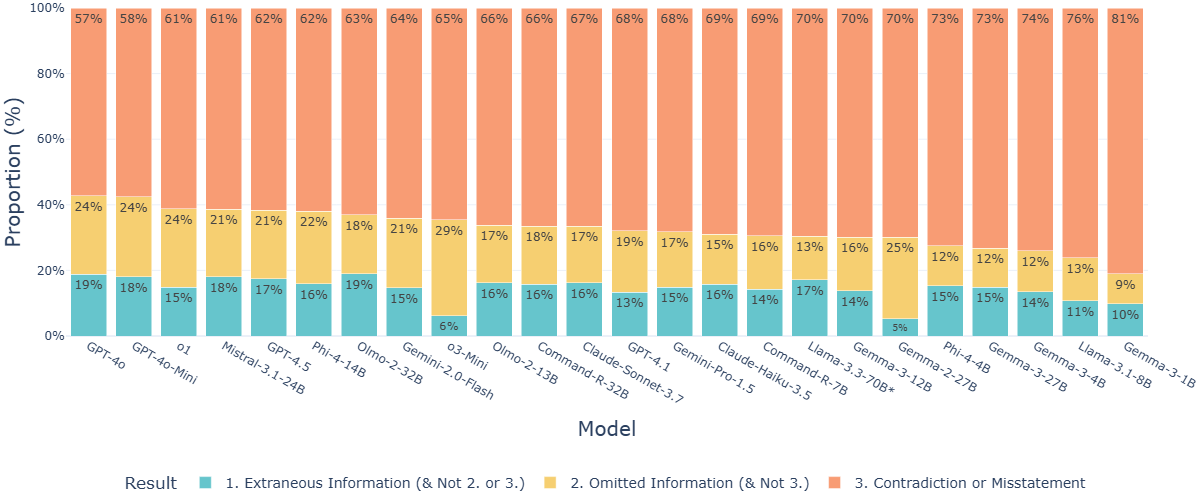} 
    \caption{Comparison of PubHealthBench-FreeForm error types classified using Llama-3.3-70bn.} \label{fig:error_props}
\end{figure}

To understand the categories of errors commonly found in found in free form responses we label the errors in the incorrect LLM responses using Llama-3.3-70bn into the following three types (Figure \ref{fig:error_props}):

\begin{enumerate}
    \item The answer adds extraneous information not in the official guidance, without contradicting it.
    \item The answer omits required points from the official guidance (i.e., is missing information).
    \item The answer contradicts or misstates the official guidance (i.e., is an incorrect deviation).
\end{enumerate}

\subsubsection{Problematic response examples}\label{supp:harm_exp}

To better understand how even the highest performing LLM responses deviate from the official guidance and the potential types of problem this could cause we also manually reviewed a subset of the errors labeled as category 3. (Section \ref{sec:error_analysis}) for the top performing model o1.

We identified a number of examples of potentially problematic deviations, which primarily involved suggesting interventions (clinical and public health) too early, too late, or that are not recommended in the guidance. To illustrate the types of mistakes identified and the issues involved we display examples of four different types of problematic response below along with the relevant official guidance passage.

\begin{figure}[H]
    \centering
    \includegraphics[width=\linewidth]{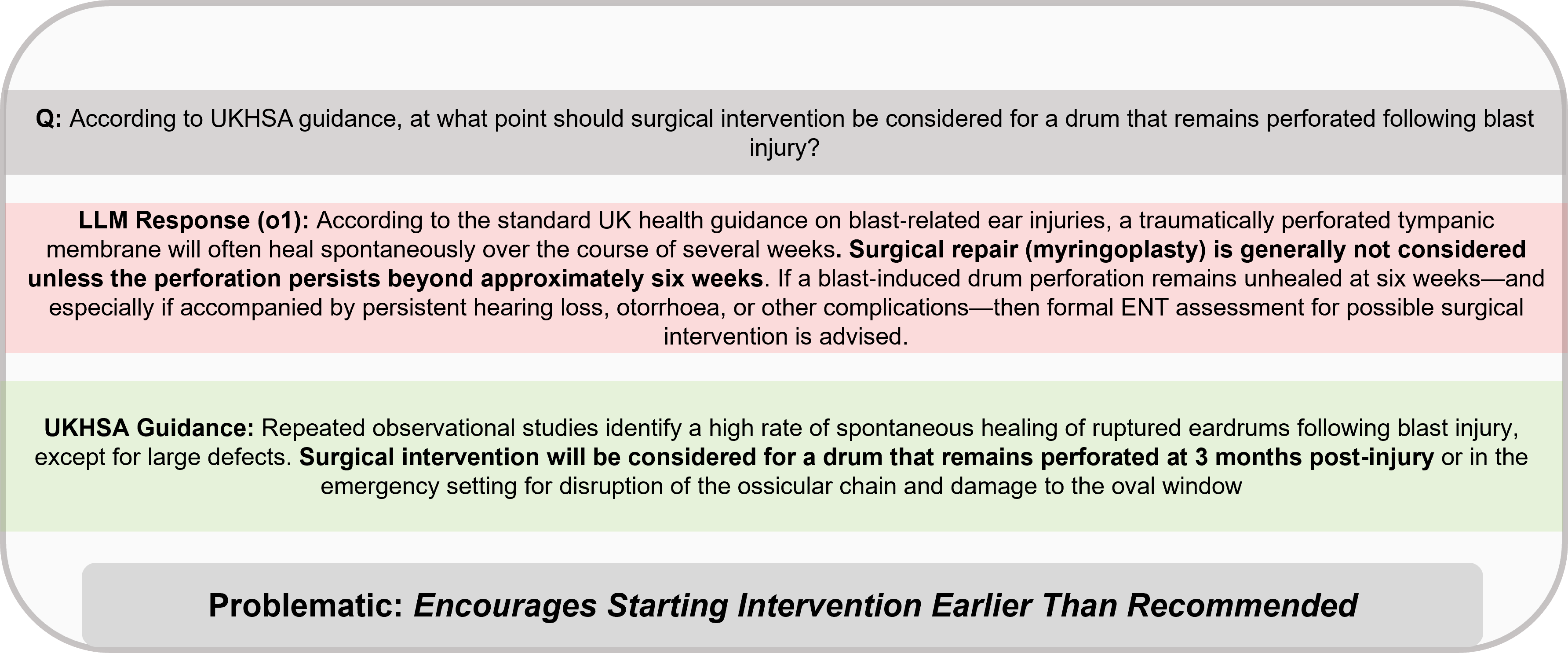} 
    \caption{Example of o1 potentially problematic responses to PubHealthBench-FreeForm.} \label{fig:ear}
\end{figure}

\begin{figure}[H]
    \centering
    \includegraphics[width=\linewidth]{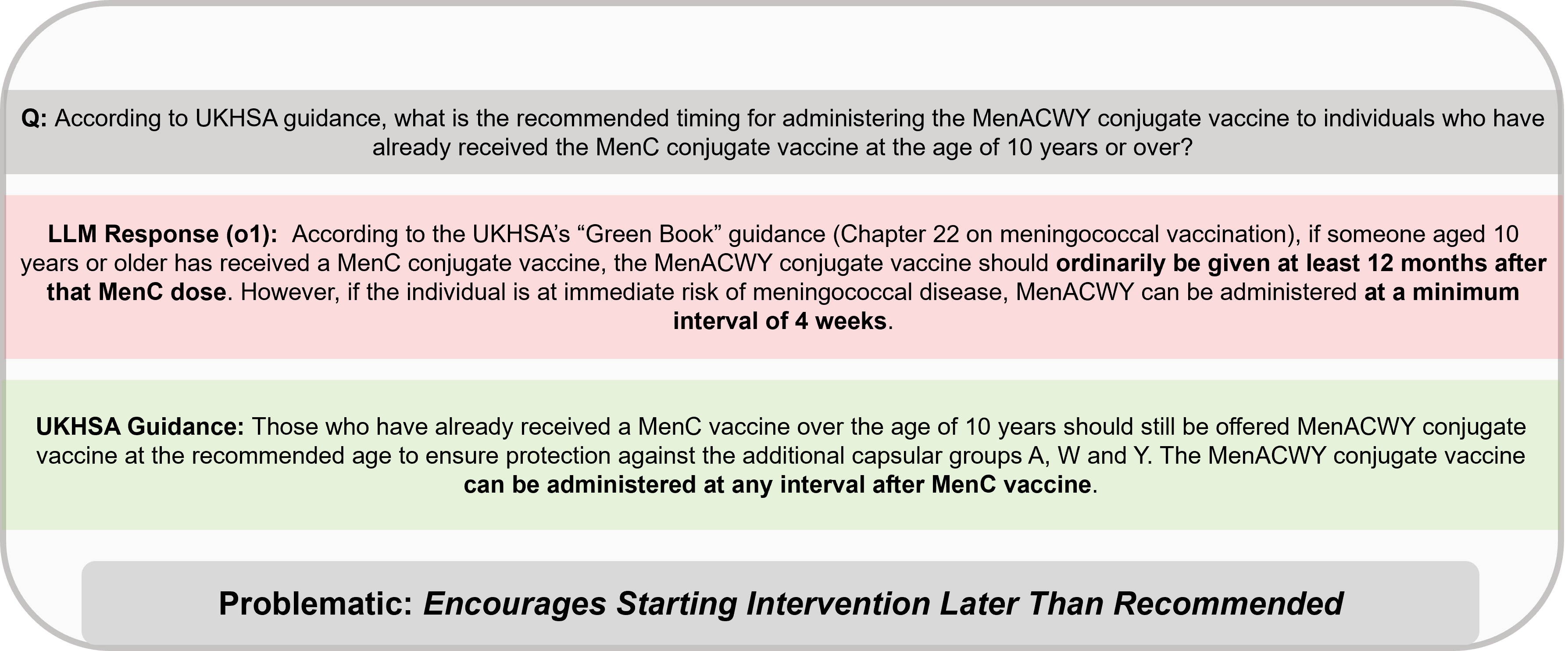} 
    \caption{Example of o1 potentially problematic responses to PubHealthBench-FreeForm.} \label{fig:menc}
\end{figure}

\begin{figure}[H]
    \centering
    \includegraphics[width=\linewidth]{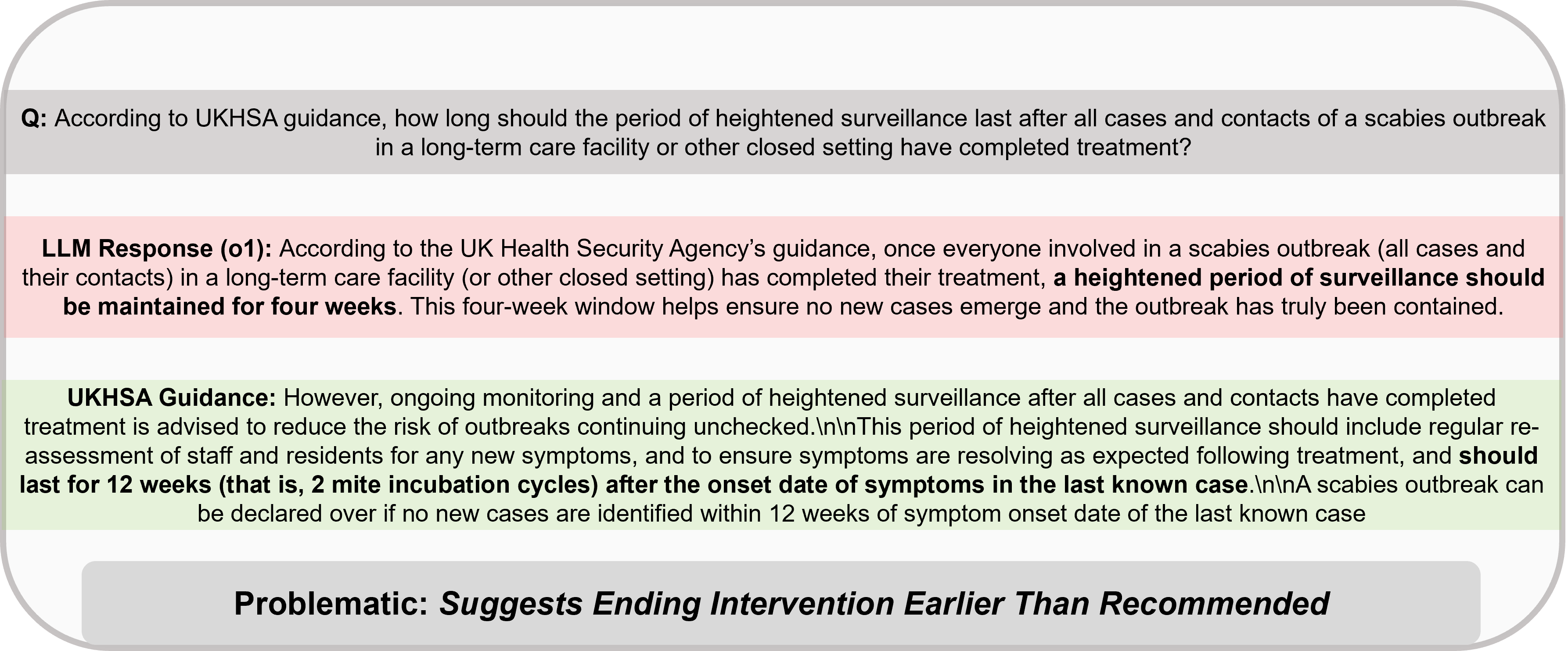} 
    \caption{Example of o1 potentially problematic responses to PubHealthBench-FreeForm.} \label{fig:scab}
\end{figure}

\begin{figure}[H]
    \centering
    \includegraphics[width=\linewidth]{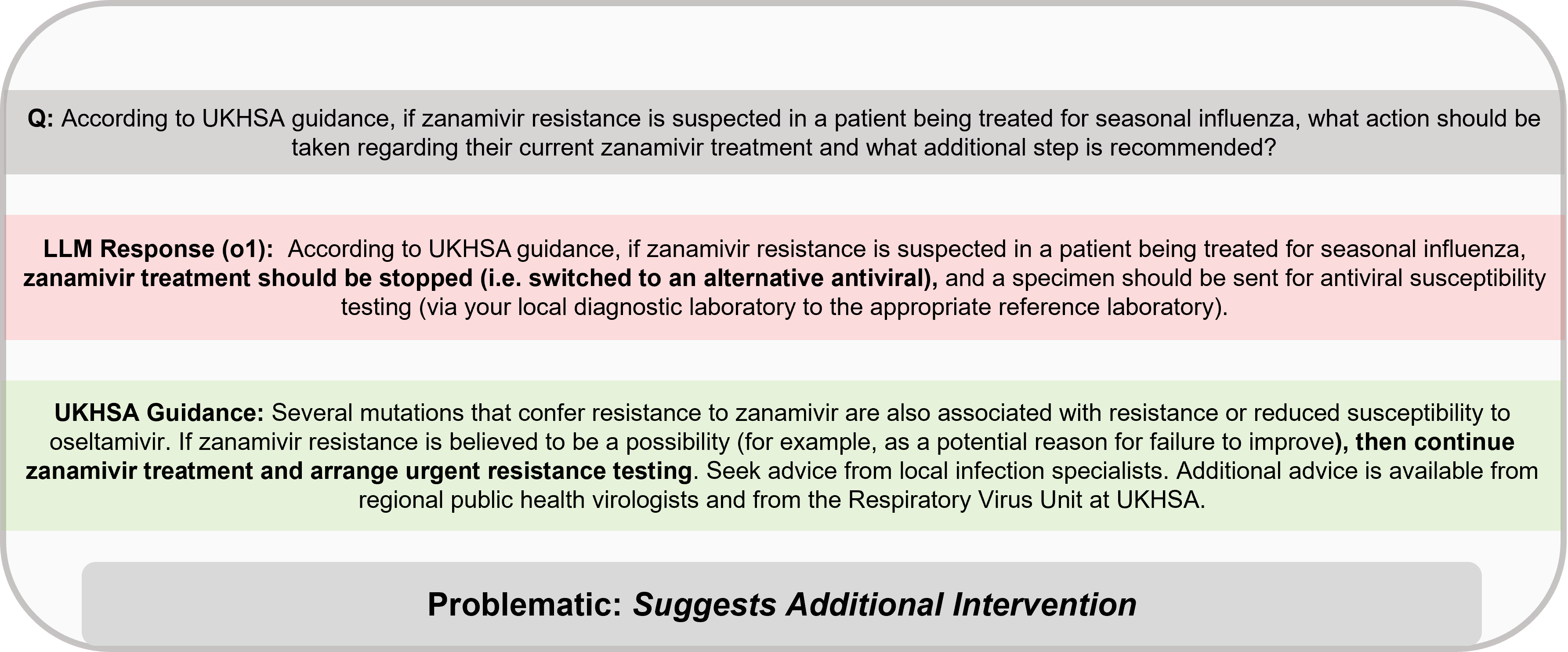} 
    \caption{Example of o1 potentially problematic responses to PubHealthBench-FreeForm.} \label{fig:viral}
\end{figure}

\subsection{Confidence intervals}\label{supp:conf_int}

\begin{table}[H]
    \caption{PubHealthBench-Full - zero-shot accuracy Wilson score 95\% confidence intervals~\citep{bowyer2025positiondontuseclt} for test set of 7929 questions, refusals included as incorrect responses, and bold indicates the highest score. *LLM used to generate benchmark.}
    \vspace{1em} 
    \label{tab:results_full_ci}
    \centering
        \resizebox{\linewidth}{!}{%
    
\begin{tabular}{lccccccccccc}
\toprule
 & \makecell{Blood Safety,\\Hepatitis, STIs\\and HIV} & \makecell{Chemicals and\\Toxicology} & \makecell{Climate and\\Health} & \makecell{Gastro and\\Food Safety} & \makecell{HCAI, Fungal,\\ AMR,\\Antimicrobial\\Use and Sepsis} & \makecell{Health Protection\\in Inclusion\\Health Settings} & \makecell{Other} & \makecell{Radiation} & \makecell{Tuberculosis,\\Travel,\\Zoonotic, and\\ Emerging infections} & \makecell{VPDs and\\Immunisation} & \makecell{Overall} \\
Model Name &  &  &  &  &  &  &  &  &  &  &  \\
\midrule
GPT-4.5 & \textbf{(87.8-93.2)} & \textbf{(89.5-92.5)} & \textbf{(95.2-98.6)} & \textbf{(87.1-92.7)} & \textbf{(92.2-96.4)} & \textbf{(92.4-95.2)} & \textbf{(87.5-94.0)} & \textbf{(85.9-92.9)} & \textbf{(90.1-93.1)} & \textbf{(91.8-94.1)} & \textbf{(91.9-93.1)} \\
o3-Mini & (84.3-90.4) & (86.7-90.1) & (91.7-96.4) & (84.7-90.9) & (89.6-94.5) & (88.8-92.2) & (83.1-90.8) & (82.7-90.5) & (85.2-88.9) & (86.7-89.6) & (88.2-89.5) \\
Gemini-2.0-Flash & (81.0-87.8) & (84.1-87.8) & (92.6-97.0) & (82.3-89.0) & (84.8-90.8) & (88.8-92.2) & (82.7-90.5) & (84.3-91.7) & (84.3-88.1) & (85.8-88.8) & (86.9-88.4) \\
Llama-3.3-70B* & (83.0-89.4) & (84.7-88.4) & (89.3-94.6) & (81.8-88.6) & (84.6-90.6) & (89.1-92.4) & (81.6-89.6) & (83.1-90.8) & (83.3-87.1) & (85.6-88.6) & (86.7-88.1) \\
Phi-4-14B & (81.8-88.4) & (80.5-84.6) & (89.0-94.4) & (81.8-88.6) & (87.3-92.8) & (86.7-90.4) & (85.5-92.6) & (80.0-88.4) & (83.0-86.9) & (83.7-86.9) & (85.3-86.8) \\
Gemini-Pro-1.5 & (77.4-84.7) & (79.5-83.6) & (91.1-95.9) & (80.3-87.3) & (83.9-90.0) & (88.1-91.6) & (81.6-89.6) & (80.4-88.7) & (82.1-86.0) & (84.4-87.5) & (84.8-86.3) \\
Mistral-3.1-24B & (81.8-88.4) & (80.6-84.6) & (89.0-94.4) & (81.1-87.9) & (86.6-92.2) & (86.6-90.3) & (83.5-91.1) & (80.0-88.4) & (80.8-84.9) & (81.7-85.1) & (84.3-85.9) \\
GPT-4o-Mini & (79.3-86.3) & (77.9-82.2) & (88.4-94.0) & (78.0-85.3) & (85.3-91.2) & (86.1-89.9) & (82.0-89.9) & (78.1-86.8) & (77.6-82.0) & (81.0-84.4) & (82.7-84.3) \\
Claude-Haiku-3.5 & (78.6-85.7) & (78.3-82.5) & (89.3-94.6) & (76.2-83.8) & (82.7-89.1) & (85.6-89.4) & (82.0-89.9) & (81.6-89.6) & (78.8-83.1) & (79.7-83.2) & (82.4-84.0) \\
Gemma-3-27B & (80.8-87.6) & (77.5-81.8) & (88.4-94.0) & (76.2-83.8) & (80.2-87.0) & (85.4-89.2) & (79.7-88.1) & (75.5-84.7) & (77.9-82.2) & (80.2-83.6) & (81.9-83.5) \\
Gemma-2-27B & (80.5-87.4) & (77.1-81.4) & (88.1-93.7) & (79.0-86.2) & (80.2-87.0) & (84.8-88.7) & (79.7-88.1) & (75.5-84.7) & (78.2-82.5) & (79.4-83.0) & (81.7-83.4) \\
Phi-4-4B & (78.1-85.3) & (77.5-81.8) & (86.9-92.9) & (77.2-84.7) & (83.6-89.8) & (83.5-87.6) & (82.0-89.9) & (72.2-81.8) & (76.3-80.7) & (78.4-82.0) & (80.9-82.6) \\
Gemma-3-12B & (76.4-83.8) & (76.7-81.1) & (86.0-92.2) & (72.2-80.3) & (79.3-86.2) & (82.2-86.4) & (82.0-89.9) & (76.3-85.3) & (76.8-81.2) & (77.3-81.0) & (80.0-81.7) \\
Command-R-32B & (75.7-83.2) & (74.9-79.4) & (86.0-92.2) & (72.5-80.5) & (80.0-86.8) & (82.6-86.8) & (78.9-87.5) & (72.5-82.1) & (76.8-81.2) & (78.8-82.4) & (79.8-81.6) \\
GPT-4o & (75.4-83.0) & (75.1-79.6) & (85.7-91.9) & (75.0-82.7) & (76.9-84.2) & (83.8-87.8) & (83.1-90.8) & (74.8-84.0) & (77.3-81.7) & (74.7-78.6) & (79.3-81.1) \\
Llama-3.1-8B & (75.2-82.8) & (74.9-79.4) & (85.7-91.9) & (73.7-81.6) & (80.5-87.2) & (82.6-86.8) & (79.7-88.1) & (72.9-82.4) & (73.8-78.4) & (77.4-81.1) & (79.1-80.9) \\
Olmo-2-32B & (72.5-80.4) & (72.3-76.9) & (85.4-91.7) & (71.2-79.4) & (79.0-86.0) & (81.1-85.4) & (75.9-85.0) & (72.2-81.8) & (74.1-78.7) & (75.1-78.9) & (77.5-79.3) \\
Command-R-7B & (72.1-80.0) & (69.9-74.6) & (82.5-89.4) & (69.7-78.1) & (72.2-80.0) & (76.6-81.3) & (78.1-86.8) & (65.9-76.3) & (67.7-72.7) & (71.1-75.1) & (73.7-75.6) \\
Olmo-2-13B & (71.8-79.8) & (67.3-72.2) & (83.9-90.6) & (69.7-78.1) & (72.7-80.5) & (76.3-81.0) & (75.9-85.0) & (68.9-78.9) & (68.9-73.8) & (70.7-74.7) & (73.5-75.4) \\
Gemma-3-4B & (69.2-77.4) & (67.2-72.1) & (81.9-88.9) & (67.0-75.6) & (69.6-77.7) & (76.7-81.5) & (76.6-85.6) & (62.3-73.1) & (67.3-72.3) & (65.3-69.5) & (71.2-73.2) \\
Gemma-3-1B & (44.9-54.2) & (46.7-52.0) & (52.9-62.8) & (40.4-49.9) & (46.3-55.5) & (48.5-54.3) & (51.4-62.7) & (39.0-50.4) & (42.1-47.5) & (39.9-44.4) & (46.5-48.7) \\
\bottomrule
\end{tabular}

    }
\end{table}

\noindent
\begin{minipage}[t]{0.48\linewidth}
  \centering
  \captionof{table}{PubHealthBench-Full zero-shot accuracy  Wilson score 95\% confidence intervals by guidance type. *LLM used to generate benchmark.}
  \label{tab:results_full_type_ci}
  \resizebox{\linewidth}{!}{
\begin{tabular}{lcccccc}
\toprule
 & \makecell{Clinical\\Guidance} & \makecell{Multiple\\Audiences} & \makecell{Professional\\Guidance} & \makecell{Public\\Guidance} & \makecell{Unclassified} & \makecell{Overall} \\
Model Name &  &  &  &  &  &  \\
\midrule
GPT-4.5 & \textbf{(90.1-92.8)} & \textbf{(91.5-95.3)} & \textbf{(91.0-92.7)} & \textbf{(94.7-97.1)} & \textbf{(90.2-93.6)} & \textbf{(91.9-93.1)} \\
o3-Mini & (84.1-87.4) & (89.3-93.6) & (87.6-89.7) & (91.0-94.2) & (86.8-90.7) & (88.2-89.5) \\
Gemini-2.0-Flash & (83.0-86.4) & (87.1-91.8) & (86.4-88.6) & (91.3-94.5) & (84.1-88.3) & (86.9-88.4) \\
Llama-3.3-70B* & (83.1-86.6) & (86.9-91.6) & (86.0-88.2) & (89.8-93.3) & (84.9-89.1) & (86.7-88.1) \\
Phi-4-14B & (82.7-86.2) & (85.4-90.4) & (84.3-86.6) & (88.5-92.2) & (82.7-87.1) & (85.3-86.8) \\
Gemini-Pro-1.5 & (80.6-84.2) & (87.4-92.1) & (83.6-85.9) & (88.6-92.3) & (83.6-88.0) & (84.8-86.3) \\
Mistral-3.1-24B & (79.5-83.2) & (84.4-89.5) & (83.7-86.1) & (88.0-91.8) & (83.7-88.0) & (84.3-85.9) \\
GPT-4o-Mini & (78.7-82.5) & (82.9-88.3) & (81.8-84.3) & (85.7-89.8) & (81.5-86.1) & (82.7-84.3) \\
Claude-Haiku-3.5 & (77.5-81.4) & (82.4-87.9) & (81.7-84.2) & (85.2-89.3) & (82.6-87.0) & (82.4-84.0) \\
Gemma-3-27B & (76.7-80.6) & (83.1-88.4) & (81.0-83.5) & (85.6-89.7) & (81.5-86.1) & (81.9-83.5) \\
Gemma-2-27B & (77.4-81.3) & (81.4-87.0) & (80.9-83.4) & (85.3-89.4) & (81.0-85.6) & (81.7-83.4) \\
Phi-4-4B & (76.2-80.2) & (81.4-87.0) & (80.5-83.0) & (83.1-87.5) & (80.0-84.7) & (80.9-82.6) \\
Gemma-3-12B & (74.9-78.9) & (80.1-85.9) & (79.4-82.0) & (84.8-89.1) & (77.6-82.6) & (80.0-81.7) \\
Command-R-32B & (75.7-79.7) & (80.1-85.9) & (78.2-80.8) & (84.6-88.9) & (80.1-84.8) & (79.8-81.6) \\
GPT-4o & (71.5-75.8) & (80.3-86.0) & (79.6-82.2) & (83.8-88.2) & (78.1-83.0) & (79.3-81.1) \\
Llama-3.1-8B & (74.7-78.8) & (79.1-85.0) & (78.8-81.4) & (80.1-84.9) & (78.6-83.4) & (79.1-80.9) \\
Olmo-2-32B & (72.4-76.6) & (78.6-84.6) & (76.7-79.4) & (82.5-87.0) & (74.7-79.9) & (77.5-79.3) \\
Command-R-7B & (67.8-72.2) & (70.1-76.9) & (74.2-76.9) & (76.7-81.8) & (72.6-78.0) & (73.7-75.6) \\
Olmo-2-13B & (67.3-71.8) & (73.8-80.2) & (72.8-75.6) & (78.6-83.5) & (72.1-77.5) & (73.5-75.4) \\
Gemma-3-4B & (62.3-66.9) & (70.7-77.4) & (71.7-74.6) & (75.0-80.2) & (71.9-77.3) & (71.2-73.2) \\
Gemma-3-1B & (37.8-42.5) & (41.4-49.1) & (48.7-52.0) & (45.5-51.8) & (47.3-53.6) & (46.5-48.7) \\
\bottomrule
\end{tabular}}
  \vspace{1em}
\end{minipage}\hfill
\begin{minipage}[t]{0.48\linewidth}
  \centering
  \captionof{table}{PubHealthBench-Reviewed Wilson score 95\% confidence intervals zero-shot accuracy by question and response type. *LLM used to generate benchmark, **Headline result.}
  \label{tab:results_verified_ci}
  \resizebox{\linewidth}{!}{
\begin{tabular}{lcccc}
\toprule
 & \makecell{Exc. Refusals} & \makecell{Inc. Refusals**} & \makecell{Invalid MCQA} & \makecell{Valid MCQA} \\
Model Name &  &  &  &  \\
\midrule
GPT-4.5 & \textbf{(90.8-94.5)} & \textbf{(90.8-94.5)} & (56.4-82.8) & \textbf{(92.2-95.6)} \\
GPT-4.1 & (90.1-93.9) & (90.1-93.9) & \textbf{(64.1-88.3)} & (90.9-94.7) \\
o1 & (89.7-93.6) & (89.7-93.6) & (51.6-79.0) & (91.2-94.9) \\
Gemini-2.0-Flash & (86.1-90.6) & (86.0-90.5) & (46.8-75.0) & (87.6-92.0) \\
o3-Mini & (85.8-90.4) & (85.8-90.4) & (54.0-80.9) & (87.0-91.5) \\
Claude-Sonnet-3.7 & (90.2-94.1) & (85.2-89.9) & (44.5-73.0) & (87.0-91.5) \\
Llama-3.3-70B* & (84.8-89.5) & (84.8-89.5) & (46.8-75.0) & (86.3-91.0) \\
Phi-4-14B & (84.3-89.1) & (84.3-89.1) & (51.6-79.0) & (85.4-90.2) \\
Gemini-Pro-1.5 & (83.5-88.5) & (83.5-88.5) & (44.5-73.0) & (85.1-89.9) \\
Mistral-3.1-24B & (82.0-87.1) & (82.0-87.1) & (46.8-75.0) & (83.3-88.4) \\
GPT-4o-Mini & (81.2-86.4) & (81.2-86.4) & (37.7-66.6) & (83.0-88.2) \\
Claude-Haiku-3.5 & (80.5-85.8) & (80.5-85.8) & (42.2-70.9) & (82.0-87.3) \\
Gemma-3-27B & (80.1-85.4) & (80.1-85.4) & (44.5-73.0) & (81.4-86.7) \\
Gemma-2-27B & (80.1-85.4) & (80.1-85.4) & (39.9-68.8) & (81.7-87.0) \\
Phi-4-4B & (78.8-84.3) & (78.8-84.3) & (49.2-77.0) & (79.8-85.3) \\
Llama-3.1-8B & (78.1-83.7) & (78.1-83.7) & (42.2-70.9) & (79.5-85.1) \\
Command-R-32B & (77.8-83.4) & (77.8-83.4) & (44.5-73.0) & (79.1-84.7) \\
GPT-4o & (89.4-93.6) & (77.7-83.3) & (37.7-66.6) & (79.4-84.9) \\
Gemma-3-12B & (77.3-82.9) & (77.3-82.9) & (46.8-75.0) & (78.3-84.0) \\
Olmo-2-32B & (75.1-80.9) & (75.1-80.9) & (39.9-68.8) & (76.4-82.3) \\
Olmo-2-13B & (71.9-78.1) & (71.9-78.1) & (42.2-70.9) & (72.9-79.2) \\
Gemma-3-4B & (70.2-76.4) & (70.2-76.4) & (39.9-68.8) & (71.2-77.6) \\
Command-R-7B & (69.6-75.9) & (69.6-75.9) & (42.2-70.9) & (70.5-76.9) \\
Gemma-3-1B & (42.4-49.5) & (42.4-49.5) & (15.3-41.1) & (43.4-50.7) \\
\bottomrule
\end{tabular}
}
\end{minipage}

\begin{minipage}[t]{\linewidth}
\centering
\captionof{table}{PubHealthBench-FreeForm model accuracy Wilson score 95\% confidence intervals by guidance audience. *LLM used to generate benchmark, **Judge LLM.}
  \vspace{1em} 
\label{tab:judge_results_ci}
\resizebox{\linewidth}{!}{
\begin{tabular}{lcccccc}
\toprule
 & \makecell{Clinical\\Guidance} & \makecell{Multiple\\Audiences} & \makecell{Professional\\Guidance} & \makecell{Public\\Guidance} & \makecell{Unclassified} & \makecell{Total} \\
Model Name &  &  &  &  &  &  \\
\midrule
o1 & \textbf{(64.3-77.6)} & \textbf{(69.6-89.3)} & (64.9-74.4) & \textbf{(76.5-91.9)} & \textbf{(72.7-88.3)} & \textbf{(71.0-77.2)} \\
GPT-4.1 & (57.8-71.8) & (58.6-81.2) & \textbf{(66.1-75.5)} & (72.1-89.0) & (58.8-77.3) & (67.2-73.7) \\
o3-Mini & (57.8-71.8) & (65.9-86.6) & (63.7-73.3) & (67.8-85.9) & (64.5-82.0) & (66.9-73.4) \\
GPT-4o & (52.6-67.0) & (58.6-81.2) & (49.2-59.5) & (72.1-89.0) & (53.3-72.5) & (57.4-64.3) \\
GPT-4.5 & (51.5-65.9) & (58.6-81.2) & (48.3-58.6) & (66.4-84.9) & (49.0-68.6) & (55.7-62.6) \\
Claude-Sonnet-3.7 & (49.7-64.2) & (53.4-76.9) & (49.7-60.0) & (65.1-83.8) & (46.8-66.6) & (55.1-62.1) \\
Gemini-2.0-Flash & (48.6-63.1) & (51.7-75.4) & (47.5-57.8) & (66.4-84.9) & (51.1-70.6) & (54.4-61.4) \\
Gemma-3-27B & (42.4-57.0) & (48.3-72.4) & (47.8-58.1) & (62.3-81.7) & (41.6-61.5) & (51.2-58.2) \\
Gemini-Pro-1.5 & (38.5-53.1) & (44.9-69.4) & (46.1-56.4) & (51.7-72.7) & (51.1-70.6) & (49.3-56.4) \\
Gemma-3-12B & (43.0-57.6) & (40.0-64.7) & (42.7-53.1) & (63.7-82.7) & (44.7-64.6) & (48.8-55.9) \\
Claude-Haiku-3.5 & (44.1-58.7) & (48.3-72.4) & (35.3-45.5) & (57.0-77.3) & (37.5-57.4) & (44.6-51.7) \\
GPT-4o-Mini** & (30.3-44.5) & (41.7-66.3) & (34.8-44.9) & (57.0-77.3) & (31.4-51.0) & (39.8-46.8) \\
Llama-3.3-70B* & (28.2-42.2) & (40.0-64.7) & (32.9-42.9) & (49.2-70.4) & (30.4-49.9) & (37.4-44.3) \\
Mistral-3.1-24B & (29.3-43.3) & (27.6-51.7) & (31.5-41.5) & (53.0-73.9) & (30.4-49.9) & (36.3-43.3) \\
Phi-4-14B & (26.1-39.8) & (33.7-58.3) & (32.6-42.6) & (47.9-69.2) & (30.4-49.9) & (36.1-43.0) \\
Olmo-2-32B & (30.3-44.5) & (38.4-63.2) & (31.8-41.7) & (44.1-65.7) & (20.8-38.9) & (35.4-42.3) \\
Gemma-3-4B & (19.3-32.1) & (27.6-51.7) & (30.7-40.6) & (46.6-68.0) & (34.4-54.2) & (33.4-40.2) \\
Command-R-32B & (23.4-36.9) & (29.1-53.4) & (24.5-33.9) & (41.6-63.3) & (32.4-52.1) & (30.8-37.5) \\
Olmo-2-13B & (24.0-37.5) & (21.7-44.9) & (25.3-34.8) & (49.2-70.4) & (25.5-44.5) & (30.5-37.3) \\
Gemma-2-27B & (20.8-33.9) & (24.6-48.3) & (27.2-36.8) & (37.9-59.6) & (25.5-44.5) & (29.8-36.4) \\
Command-R-7B & (14.7-26.5) & (9.5-28.5) & (18.0-26.6) & (35.5-57.1) & (12.6-28.5) & (20.4-26.4) \\
Llama-3.1-8B & (11.3-22.2) & (13.4-34.1) & (12.1-19.6) & (28.4-49.6) & (11.7-27.3) & (16.2-21.7) \\
Phi-4-4B & (11.3-22.2) & (14.7-36.0) & (13.9-21.8) & (20.5-40.4) & (9.2-23.7) & (15.9-21.5) \\
Gemma-3-1B & (9.4-19.6) & (5.9-22.5) & (16.9-25.4) & (17.3-36.3) & (8.4-22.5) & (15.6-21.1) \\
\bottomrule
\end{tabular}

}
\end{minipage}\hfill

\end{document}